\documentclass[11pt]{article}
\usepackage[utf8]{inputenc}
\usepackage[top=1in, left=1in, right=1in, bottom=1in]{geometry}

\usepackage{bm,amsmath,amsthm,amssymb,multicol,algorithmic,algorithm,enumitem,graphicx,subfigure}
\usepackage{microtype}
\usepackage{graphicx}
\usepackage{subfigure}
\usepackage{booktabs}
\usepackage{natbib}
\setcitestyle{open={(},close={)}}

\usepackage{mathtools}

\DeclarePairedDelimiter{\floor}{\lfloor}{\rfloor}

\newtheorem{theorem}{Theorem}[section]
\newtheorem{corollary}[theorem]{Corollary}
\newtheorem{lemma}[theorem]{Lemma}

\usepackage[colorlinks=true,citecolor=blue]{hyperref}

\begin{document}

\title{Toward Communication Efficient Adaptive Gradient Method}

\author{\textbf{Xiangyi Chen, Xiaoyun Li, Ping Li} \\\\
Cognitive Computing Lab\\
Baidu Research\\
10900 NE 8th St. Bellevue, WA 98004, USA\\
  \texttt{\{xiangyichen1900,lixiaoyun996,pingli98\}@gmail.com}
}

\date{\vspace{0.5in}}
\maketitle

\begin{abstract}\vspace{0.1in}
\noindent\footnote{The work was conducted in Summer 2019, while Xiangyi Chen and Xiaoyun Li were research  interns at Baidu Research -- Bellevue WA.}In  recent years, distributed optimization is proven to be an effective approach to accelerate training of large scale machine learning models such as deep neural networks. With the increasing computation power of GPUs, the bottleneck of training speed in distributed training is gradually shifting from computation to communication. Meanwhile, in the hope of training machine learning models on mobile devices, a new distributed training paradigm called ``federated learning'' has become popular. The communication time in federated learning is especially important due to the low bandwidth of mobile devices. While various approaches to improve the communication efficiency have been proposed for federated learning, most of them are designed with SGD as the prototype training algorithm. While adaptive gradient methods have been proven effective for training neural nets, the study of adaptive gradient methods in federated learning is scarce.  In this paper, we propose an adaptive gradient method that can guarantee both the convergence and the communication efficiency for federated learning.
\end{abstract}

\textbf{Keywords:}\  {Federated Learning, Adaptive Method, Optimization, Convergence Analysis}

\maketitle

\newpage

\section{Introduction}

Distributed training has been proven to be a successful way of accelerating training large scale machine learning models, for example, training ultra-large-scale CTR (click-through rate) models in commercial search engines~\cite{Proc:Fan_KDD19,Proc:Zhao_MLSys20,Proc:Xu_SIGMOD21}. With the advances of computing power and algorithmic design, one can now train models that need to be trained for days even weeks in the past within just a few minutes~\citep{you2019large}. When the computing power is high compared with the network bandwidth connecting different machines in distributed training, the training speed can be bottlenecked by the transmission of gradients and parameters. Such situation occurs increasingly more often in recent years due to the rapid growth in power of GPUs. Therefore, reducing communication overhead is gradually becoming an important research direction in distributed training~\citep{alistarh2017qsgd,wangni2018gradient, lin2017deep}.
In addition, a new training paradigm called \textit{Federated Learning}~\citep{konevcny2016federated,mcmahan2016communication} was proposed recently, where models are trained distributively with mobile devices as workers and data holders. Consider the case where the data is stored and the model is trained on each user's mobile device (e.g.,~\cite{Proc:Wang_WWW20}). The existence of limited bandwidth necessitates the development of communication efficient training algorithms. Moreover, it is unpractical for every user to keep communicating with the central server, due to the power condition or wireless connection of the device. To cope with  communication issues, an SGD-based algorithm with periodic model averaging called \textit{Federated Averaging} is proposed in~\cite{mcmahan2016communication}.

Federated learning extends the traditional parameter server setting, where the data are located on different workers and information is aggregated at a central parameter server to coordinate the training. In the parameter server setting, many effective communication reduction techniques were proposed such as gradient compression~\citep{lin2017deep,bernstein2018signsgd} and quantization~\citep{alistarh2017qsgd,wangni2018gradient, wen2017terngrad} for distributed SGD. In federated learning, one can substantially reduce  communication cost by avoiding frequent transmission between local workers and central server. The workers train and maintain their own models locally, and the central server aggregates and averages the model parameters of all workers periodically. After averaging, new model parameter is fed back to each local worker, which starts another round of ``local training + global averaging''. Some of the aforementioned communication reduction techniques can also be incorporated into Federated Averaging to further reduce the communication cost~\citep{reisizadeh2019fedpaq}.

Despite the active efforts on improving algorithms based on periodic model averaging~\citep{haddadpour2019trading,reisizadeh2019fedpaq,FedSKETCH2020}, the prototype algorithm is still SGD. On the other hand, we know that adaptive gradient methods such as AdaGrad~\citep{duchi2011adaptive}, Adam~\citep{kingma2014adam} and AMSGrad~\citep{reddi2019convergence} often perform better than SGD when training neural nets, in terms of difficulty of parameter tuning and convergence speed at early stages. This motivates us to study adaptive gradient methods in federated~learning.

\vspace{0.15in}
\noindent
\textbf{Our contributions.}\hspace{0.05in}We study how to incorporate adaptive gradient method into federated learning. Specifically, we show that unlike SGD, a naive combination of adaptive gradient methods and periodic model averaging results in algorithms that may fail to converge. Based on federated averaging and decentralized training, we propose an adaptive gradient method with communication cost sublinear in $T$ that is guaranteed to converge to stationary points with a rate of $O(\sqrt{d}/\sqrt{TN})$, where $d$ is the dimension of the problem, $T$ is the number of iterations, and $N$ is the number of workers (nodes). Our proposed method enjoys the benefit from both worlds: the fast convergence performance of adaptive gradient methods and low communication cost of federated learning.

\section{Related Work}

\textbf{Federated learning.}
A classical framework for distributed training is the parameter server framework. In such a setting, a parameter server is used to coordinate training and the main computation (e.g., gradient computation) is offloaded to workers in parallel. For SGD under this framework~\citep{recht2011hogwild,li2014scaling, zinkevich2010parallelized,Proc:Zhao_MLSys20}, the gradients can be computed by workers on their local data and sent to the parameter server which will aggregate the gradients and update the model parameters.  Recently, a variant of the parameter server setting called federated learning~\citep{mcmahan2016communication, konevcny2016federated} draws great attention. One of the key features of
federated learning is that workers are likely to be mobile devices which share a low bandwidth with the parameter server. Thus, communication cost plays a more important role in federated learning compared with the traditional parameter server setting. To reduce the communication cost,~\citet{mcmahan2016communication} proposed an algorithm called Federated Averaging which is a version of parallel SGD with local updates. In Federated Averaging, each worker updates their own model parameters locally using SGD, and the local models are synchronized by periodic averaging through the parameter server. The algorithm is also called local SGD or K-step SGD in some other papers~\citep{yu2019linear, stich2018local, zhou2017convergence}. Theoretically, it is proven in~\cite{yu2019linear} that local SGD can save a communication factor of $O(T^{1/4})$ while achieving the same convergence rate as vanilla SGD.

\vspace{0.1in}
\noindent\textbf{Adaptive gradient methods.} Adaptive gradient methods usually refer to the class of gradient based optimization algorithms that adaptively update their learning rate (for each parameter coordinate) using historical gradients. Adaptive gradient methods such as Adam~\cite{kingma2014adam}, AdaGrad~\citep{duchi2011adaptive}, AdaDelta~\citep{zeiler2012adadelta} are commonly used for training deep neural networks. It has been observed empirically that in many cases adaptive methods can outperform SGD or other methods in terms of convergence speed. There are also many variants trying to improve different aspects of these algorithms, e.g.,~\cite{reddi2019convergence,keskar2017improving, luo2019adaptive, chen2018universal, agarwal2018case}. The work~\cite{reddi2019convergence} pointed out the divergence issue of Adam, and proposed AMSGrad algorithm for a fix. Moreover, increasing efforts are investigated into theoretical analysis of these algorithms~\cite{chen2018convergence, ward2019adagrad, li2018convergence, staib2019escaping, zhou2018convergence, zou2019sufficient, zou2018convergence}. {In the federated learning setting, \citet{xie2019local} proposed a variant of AdaGrad, and \citet{reddi2020adaptive} proposed a framework of adaptive gradient methods that includes variants of AdaGrad, Adam, and Yogi~\cite{zaheer2018adaptive}, along with a convergence for the framework. There are also a few recent works trying to apply adaptive gradient methods in distributed optimization \cite{xu2020asynchronous,chen2021convergence}.
In this paper, we embed new adaptive gradient methods into federated learning and provide rigorous convergence analysis. The proposed algorithm can achieve the same convergence rate as its vanilla version, while enjoying the communication reduction brought by periodic model averaging. }

\section{Distributed Training with Periodic Model Averaging}

In this section, we  introduce our problem setting and the periodic model averaging framework for federated learning.

\vspace{0.08in}
\noindent\textbf{Notation.} Throughout the paper, $x_{t,i}$ denotes model parameter $x$ at node $i$ and iteration $t$. $a/b$ is element-wise division when $a$ and $b$ are vectors of the same dimension, and $a \cdot b$ and $a^p$ denote element-wise multiplication and power, respectively. $(a)_j$ denotes the $j$-th coordinate of vector $a$.

\subsection{Distributed Optimization }
In this paper, we consider the following formulation for distributed training, with $N$ works (nodes):
\begin{align}
\min_{x} \frac{1}{N}\sum_{i=1}^N f_i(x) \nonumber
\end{align}
where $f_i$ can be considered as the averaged loss over data at worker $i$ and the function can only be accessed by the worker itself. For instance, for training neural nets, $f_i$ can be viewed as the average loss of data located at the $i$-th node.

We consider the case where $f_i$'s might be nonconvex (e.g., deep nets). Our convergence analysis needs the following assumptions.

\vspace{0.1in}

\noindent\textbf{Assumptions:}

\textbf{A1:} Lipschitz property, $f_i$ is differentiable and $L$-smooth, i.e., $\|\nabla f_i(x) - \nabla f_i(y)\| \leq L\|x-y\|,\ \forall x,y$.

\textbf{A2:} Unbiased gradient estimator, $\mathbb E [g_{t,i}] = \nabla f_i(x_{t,i}) $.

\textbf{A3:} Coordinate-wise bounded variance for the gradient estimator $\mathbb E [(g_{t,i})_j^2] \leq \sigma^2, \forall j \in [d] $.

\textbf{A4:}
Bounded gradient estimator, $\|g_{t,i}\|_{\infty} \leq G_{\infty}$.

\vspace{0.05in}
The assumptions A1, A2, A3 are standard in stochastic optimization. A4 is a little stronger than bounded variance assumption (A2), and is commonly used in analysis for adaptive gradient methods~\citep{chen2018convergence,ward2019adagrad} to simplify the convergence analysis by bounding possible adaptive learning rates.

\subsection{Periodic Model Averaging}
Recently, a new trend in algorithm design for distributed training is using periodic model averaging to reduce communication cost. This is motivated by the fact that in some circumstances where distributed optimization is used, the computation time is dominated by communication time. This phenomenon is significantly exacerbated in federated learning, where the  bandwidth is relatively small (e.g. wireless networks on mobile devices).

Local SGD (i.e., Federated Averaging~\citep{konevcny2016federated, zhou2017convergence}) is featured by the use of periodic model averaging with SGD (see Algorithm~\ref{alg: local_sgd}). Periodic model averaging can reduce the number of communication rounds and it is shown in~\cite{yu2019linear,stich2018local} that by using periodic model averaging, one can achieve the same convergence rate as distributed SGD with a communication cost sublinear in $T$. Note that, in practical scenarios, samples distribution for each node may not be i.i.d.. In the example of training on mobile devices, if the data on each device is collected from each single user, samples on a node will no longer be randomly drawn from the population.

\begin{algorithm}[h]
	\caption{Local SGD (with N nodes)}
	\label{alg:mediangd}
	\begin{algorithmic}[1]
		\STATE {\bfseries Input:} learning rate $\alpha$, current point $x_t$ 
		\STATE  $g_{t,i}  \leftarrow \nabla f_i(x_{t,i}) + \xi_{t,i}$
		\IF{$t \mod k \neq 0$}
		\STATE $x_{t+1,i} \leftarrow x_{t,i} - \alpha g_{t,i} $
		\ELSE
		\STATE $x_{t+1,i} \leftarrow \frac{1}{N}\sum_{j=1}^{N}\left(x_{t,j} - \alpha  g_{t,i}\right)$
		\ENDIF
	\end{algorithmic}
	\label{alg: local_sgd}
\end{algorithm}

Since local SGD is heavily used for training neural nets in federated learning, it is natural to consider using adaptive gradient methods in such setting to integrate advantageous aspects of adaptive gradient methods. In the remaining sections of this paper, we will study how to use periodic model averaging with adaptive gradient methods.

\section{Adaptive gradient methods with periodic model averaging}
In this section, we  explore the possibilities of combining adaptive gradient method with periodic model averaging. We  use  AMSGrad~\citep{reddi2019convergence} as our prototype algorithm due to its nice convergence guarantee and superior empirical performance. The proposed scheme will be called local AMSGrad.

\subsection{Divergence of naive local AMSGrad}

Similar to local SGD, the most straightforward way to combine AMSGrad with periodic model averaging works as follows:

1. Each node runs AMSGrad locally.

2. The variables $\{x_{t,i}\}_{i=1}^N$ are averaged every $k$ iterations.

\vspace{0.1in}

The algorithm's pseudo code is shown in Algorithm~\ref{alg: naive_local_ams}.

\begin{algorithm}[h]
	\caption{Naive local AMSGrad (with N nodes)}
	\label{alg:mediangd}
	\begin{algorithmic}[1]
		\STATE {\bfseries Input:} learning rate $\alpha$, point $x_t$, $m_{0,i} = 0, \hat v_{0,i} = \epsilon \mathbf{1}, \forall i$
		\STATE  $g_{t,i}  \leftarrow \nabla f_i(x_{t,i}) + \xi_{t,i}$
		\STATE $m_{t,i} = \beta_1 m_{t-1,i} + (1-\beta_1) g_{t,i}$
		\STATE $v_{t,i} = \beta_2 v_{t-1,i}+(1-\beta_2)g_{t,i}^2$
		\STATE $\hat{v}_{t,i} = \max( v_{t,i},\hat{v}_{t-1,i}) $
		\IF{$t \mod k \neq 0$}
		\STATE $x_{t+1,i} \leftarrow x_{t,i} - \alpha  \frac{m_{t,i}}{\sqrt{\hat{v}_{t,i}}}$
		\ELSE
		\STATE $x_{t+1,i} \leftarrow \frac{1}{N}\sum_{j=1}^{N}\left(x_{t,j} - \alpha  \frac{m_{t,j}}{\sqrt{\hat{v}_{t,i}}}\right)$
		\ENDIF
	\end{algorithmic}
	\label{alg: naive_local_ams}
\end{algorithm}

Since Algorithm~\ref{alg: naive_local_ams} is  similar to Algorithm~\ref{alg: local_sgd} except for the use of adaptive learning rate, given that Algorithm~\ref{alg: local_sgd} is guaranteed to converge to stationary points, one may expect that Algorithm~\ref{alg: naive_local_ams} is also guaranteed to converge. However, this is not necessarily the case.  Algorithm~\ref{alg: naive_local_ams} can fail to converge to stationary points, due to the possibility that the adaptive learning rates on different nodes are different. We show this possibility in Theorem~\ref{thm: diverge}.

\begin{theorem}\label{thm: diverge}
There exists a problem where Algorithm~\ref{alg: naive_local_ams} converges to non-stationary points no matter how small the stepsize is.
\end{theorem}
\noindent\textbf{Proof}: We prove by providing a counter example. Consider a simple 1-dimensional case where $N=3$ with
\begin{equation*}
    f_1=\begin{cases}
        2x^2, &|x|\leq 1,\\
        4|x|-2, &|x|>1.
    \end{cases} \quad\quad
    f_2=f_3=\begin{cases}
        -0.5x^2, &|x|\leq 1,\\
        -|x|+0.5, &|x|>1.
    \end{cases}
\end{equation*}
It is clear that $f(x)=\sum_{i=1}^3 f_i(x)$ has a unique stationary point at $x=0$ such that $\nabla f(x)=0$. Suppose $\alpha=0.1$, $\beta_1=0$, $\beta_2=0.5$ and the initial point is $x_{0,i}=5$ for $i=1,2,3$. Also suppose that $k=1$, i.e., we average local parameters after every iteration. At $t=1$, for the first node associated with $f_1$, we have $m_{0,1}=g_{0,1}=4$, and $\hat v_{0,1}=0.5\times 4^2=8$. For $i=2,3$, we have $m_{0,i}=g_{0,i}=-1$ and $\hat v_{0,i}=0.5$. Since in the naive method every node keeps its own learning rate, after the first update we have $x_{1,1}=5-\frac{0.1\times 4}{2\sqrt 2}\approx 4.86$, $x_{1,2}=x_{1,3}=5+0.1\sqrt 2\approx 5.15$. By Algorithm~\ref{alg: naive_local_ams}, we have $x_{1}\approx 5.05$, which heads towards the opposite direction of the true stationary point. We can then continue to show that for $t>1$, $m_{t,1}=4$, $\hat v_{t,1}=(1-0.5^t)\times 4^2$ and $m_{t,i}=-1$, $\hat v_{t,i}=1-0.5^t$ for $i=2,3$. Therefore, we always update $x_{t,1}$ by $-0.1/\sqrt{1-0.5^t}$, while updating $x_{t,2}$ and $x_{t,3}$ by $0.1/\sqrt{1-0.5^t}$. As a result, the averaged model parameter will head towards $+\infty$, instead of 0. The above argument can be trivially extended to arbitrary stepsize since the gradients do not change in the linear region of the function. \hfill $\square$

\vspace{0.1in}

Given the example of divergence shown in the proof of Theorem~\ref{thm: diverge}, we know that a naive combination of periodic model averaging and adaptive gradient methods may not be valid even in a very simple case. By diving into the example where the algorithm fails, one can notice that the divergence is caused by the non-consensus of adaptive learning rates on different nodes. This suggests that we should keep the adaptive learning rate the same at different nodes. Next, we incorporate this idea into algorithm design to use shared adaptive learning rate on different nodes.

\subsection{Local AMSGrad with shared adaptive learning rates}

\begin{figure}[h]
\centering
\mbox{
		\includegraphics[width=5in]{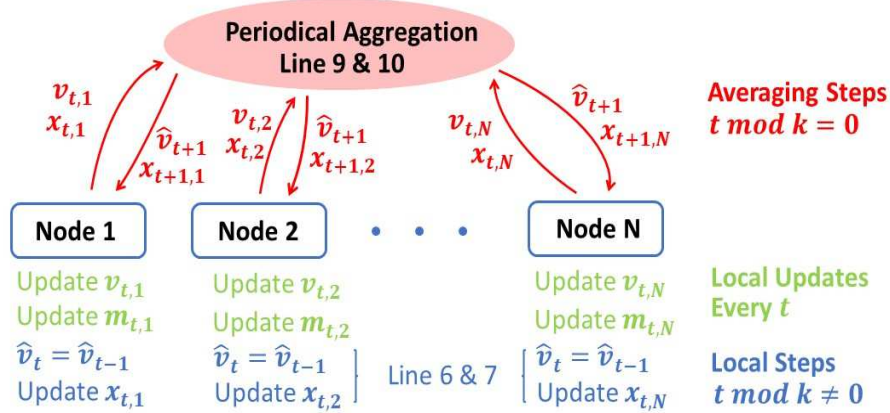}
}
	\vspace{-0.15in}
	\caption{Illustration of the proposed local AMSGrad scheme (Algorithm~\ref{alg: local_ams}) with shared adaptive learning rate. The local servers employ AMSGrad updates locally, and the global server aggregates the model parameters and second moment $\hat v$ every $k$ steps.}
	\label{fig:scheme}
\end{figure}

In the last section, we have showed an example where  a naive combination of AMSGrad and periodic model averaging may diverge. The key divergence mechanism is due to the use of different adaptive learning rates on different nodes. A natural way to improve it is to force different nodes to have the same adaptive learning rate and we instantiate this idea in Figure~\ref{fig:scheme} and Algorithm~\ref{alg: local_ams}.

\begin{algorithm}[h]
	\caption{Local AMSGrad (with N nodes)}
	\begin{algorithmic}[1]
		\STATE {\bfseries Input:} learning rate $\alpha$, point $x_t$, $m_{t,i} = 0, \hat v_{0,i} = \epsilon \mathbf{1}, \forall i$
		\STATE  $g_{t,i}  \leftarrow \nabla f_i(x_{t,i}) + \xi_{t,i}$
		\STATE $m_{t,i} = \beta_1 m_{t-1,i} + (1-\beta_1) g_{t,i}$
		\STATE $v_{t,i} = \beta_2 v_{t-1,i}+(1-\beta_2)g_{t,i}^2$
		\IF{$t \mod k \neq 0$}
		\STATE $\hat{v}_{t} = \hat{v}_{t-1}$  	
		\STATE $x_{t+1,i} \leftarrow x_{t,i} - \alpha  \frac{m_{t,i}}{\sqrt{\hat{v}_t}}$
		\ELSE
		\STATE $\hat{v}_{t} = \max(\frac{1}{N}\sum_{i=1}^N v_{t,i},\hat{v}_{t-1}) $
		\STATE $x_{t+1,i} \leftarrow \frac{1}{N}\sum_{j=1}^{N}\left(x_{t,j} - \alpha  \frac{m_{t,j}}{\sqrt{\hat{v}_t}}\right)$
		\ENDIF
	\end{algorithmic}
	\label{alg: local_ams}
\end{algorithm}

Compared with Algorithm~\ref{alg: naive_local_ams}, Algorithm~\ref{alg: local_ams} introduces a periodic averaging step for $v_{t,i}$ and updates $\hat v_t$ at the server side, the same $\hat v_t$ is used for local updates of different nodes. The intuition behind the design is that, since $v_{t,i}$ can be viewed as second moment estimation of the gradients and the average of $v_{t,i}$ is also an estimation of second moment, we expect the performance of the proposed method to be close to original AMSGrad.   Note that this is not the only way to synchronize adaptive learning rate at different nodes, e.g., one can keep $\hat v_{t,i}$ locally and use the average of $\hat v_{t,i}$ obtained at the previous averaging step as the adaptive learning rate during local updates.  With the synchronization of adaptive learning rates, the divergence example in Theorem~\ref{thm: diverge} is no longer valid. Nevertheless, the convergence guarantee of the proposed algorithm is still not clear since it uses periodic averaging with adaptive learning rates and momentum. In the next section, we will establish the convergence guarantee of the proposed algorithm.

\section{Convergence of Local AMSGrad}

In this section, we analyze the convergence behavior of Algorithm~\ref{alg: local_ams}. The main result is summarized in Theorem~\ref{thm: loca_ams}.

\begin{theorem}\label{thm: loca_ams}
For Algorithm~\ref{alg: local_ams}, if A1 - A4 are satisfied, define $\overline x_t = \frac{1}{N} \sum_{i=1} ^N x_{t,i}$, set $\alpha = \min(\frac{\sqrt{N}}{\sqrt{Td}},\frac{\sqrt{\epsilon}}{4L})$, we have for any $T\geq \frac{16NL^2}{\epsilon d}$,  \begin{align}\label{eq: local_ams}
   \frac{1}{T}  \sum_{t=1} ^T \mathbb E \left[ \left\| \frac{ {\nabla f} (\overline x_t)}{\hat v_t^{1/4}} \right\|^2\right] \leq  & 8\frac{\sqrt{d}}{\sqrt{TN}} (\mathbb E[   f(\overline  x_{1})] -  \min_x f(x)) + 8{L}  \frac{\sqrt{d}}{\sqrt{TN}}  \sigma^2  \frac{1}{\epsilon}\nonumber \\
    &   + 8\frac{d}{T}\frac{\beta_1}{1-\beta_1 } G^2   \frac{1}{\epsilon^{1/2}} +   8\frac{LN}{T^2} \frac{\beta_1^2}{(1-\beta_1)^2 } \frac{G^2 }{\epsilon} \nonumber  \\
    &+  8\frac{N}{T}   {L} \left( \frac{\beta_1^2}{(1-\beta_1)^2} +5(k-1)^2  \right)    \frac{G^2}{\epsilon^{1.5}}.
\end{align}
\end{theorem}

From Theorem~\ref{thm: loca_ams}, we can analyze the effect of different factors on the convergence rate of Algorithm~\ref{alg: local_ams}. The first two terms are standard in convergence analysis, which are introduced by initial function value and the variance of the gradient estimator. Note that the $\sqrt{d}$ factor in the two terms is due to the bounded coordinate-wise variance assumption, which makes the total variance of the gradient estimator upper bounded by $d\sigma^2$. One can remove the dependency on $d$ on these two terms by assuming bounded total variance. The terms diminishes with $\beta_1$ are introduced by the use of momentum in $m_{t,i}$. The most important term for communication efficiency is the term grows with $k$, the number of local updates. This is due to the bias on update directions introduced by local updates. It is clear that this term will not dominate RHS of~\eqref{eq: local_ams} when $k \leq O(\frac{T^{1/4}d^{1/4}}{\sqrt{N}})$. Thus, one can achieve a convergence rate of $O(\sqrt{d}/{\sqrt{TN}})$ with communication rounds sublinear in $T$. This matches the convergence rate of vanilla AMSGrad $O(\sqrt{d}/\sqrt{T})$ proven in~\cite{chen2018convergence,zhou2018convergence}. Another aspect which deserves some discussion is the dependency on $\epsilon$. One may expect the RHS to be large when $\epsilon$ is small. However, this is only true when the gradients are also small such that their $L_\infty$ norm is in the order of $\epsilon$. All the dependencies on $\epsilon$ appears in lower bounding $\hat{v}_t$. With the update rule of $\hat{v}_t$, it will quickly become at least the same order as second moment of stochastic gradients, which does not achieve the worst case. Thus, one should expect that when $\epsilon$ is really small, the worst case convergence rate is not achieved in practice. As for the convergence measure, one can easily lower bound LHS of~\eqref{thm: loca_ams} by the traditional measure $\frac{1}{T} \sum_{t=1}^T \mathbb E [\|\nabla f(\bar x_t)\|^2]$ using the fact that $\|\hat v_t\| \leq G^2$. To wrap things up, we simplify the convergence rate in Corollary~\ref{cor: rate}.

\begin{corollary}\label{cor: rate}
For Algorithm~\ref{alg: local_ams}, if A1 - A4 are satisfied, set $k \leq T^{1/4}d^{1/4}/\sqrt{N}$, for $T\geq \max(Nd,\frac{16NL^2}{\epsilon d})$ when $\beta_1 >0$ and  for $T\geq \frac{16NL^2}{\epsilon d}$ when $\beta_1 = 0$, we have
\begin{align}
   & \frac{1}{T}  \sum_{t=1} ^T \mathbb E \left[ \left\| { {\nabla f} (\overline x_t)} \right\|^2\right]
    \leq  O\left(\frac{\sqrt{d}}{\sqrt{TN}}\right). \nonumber
\end{align}
\end{corollary}
Again, the $\sqrt{d}$ factor is due to the bounded coordinate-wise variance assumption A3, one can easily remove the $d$ dependency by assuming bounded total variance. The proof of Theorem \ref{thm: loca_ams} can be found in Appendix \ref{app: proof_thm5}.

\section{Experiments}
We compare the performance of local SGD, local AMSGrad, and naive local AMSGrad on a synthetic Gaussian mixture dataset~\citep{sagun2017empirical}, the letter recogintion dataset~\citep{frey1991letter}  and the standard MNIST dataset. Experiments were  conducted using the PaddlePaddle deep learning platform.

\subsection{Gaussian mixture dataset (non-i.i.d. case)}
In the first experiment, we use the synthetic dataset (Gaussian cluster data) from~\cite{sagun2017empirical}. We use 10 isotropic 100 dimensional Gaussian distributions with different mean and same standard deviation to generate the data. The standard deviation of each dimension is 1. The mean of each cluster is generated from an isotropic Gaussian distribution with marginal $N(0,1)$ for each dimension. The labels are the corresponding indices of the cluster from which the data are drawn. The model is a neural network with 2 hidden layers with 50 nodes per layer, and the activation function is ReLU for both layers. The batch size in training is 256.  We use $N=5$ workers with each worker containing data from two classes. This assignment of data corresponds to a  non-i.i.d. distribution of data on  different nodes. The average local period $k$ is set to 10. We perform the learning rate search on a log scale, and increase the learning rate starting from 1e-6 until the algorithm diverges or the performance deteriorates significantly.  Specifically, the maximum learning rate is 1 for local SGD and 1e-2 for both local AMSGrad and naive local AMSGrad.

\begin{figure}[t]
\centering
\mbox{
	\includegraphics[width=3in]{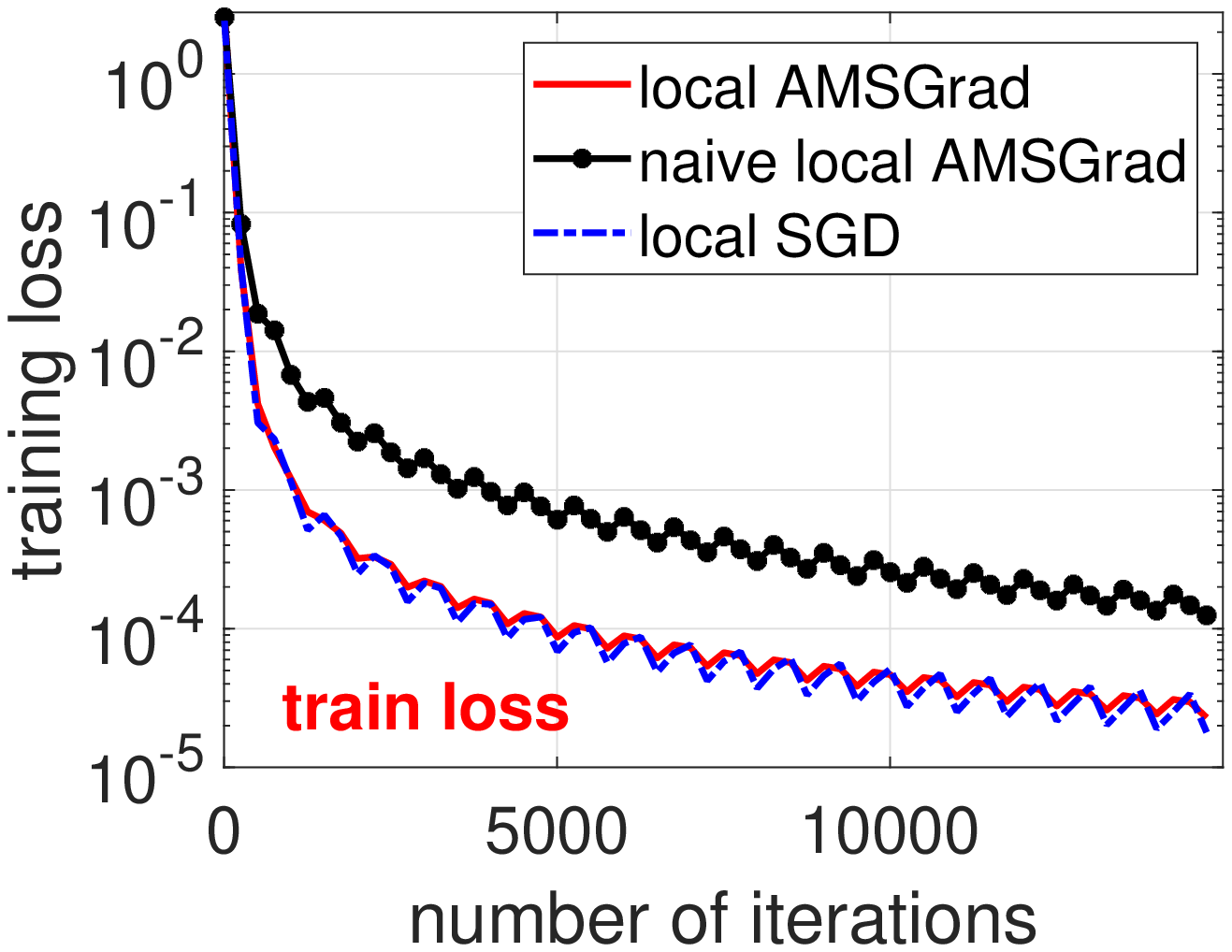}
		\includegraphics[width=3in]{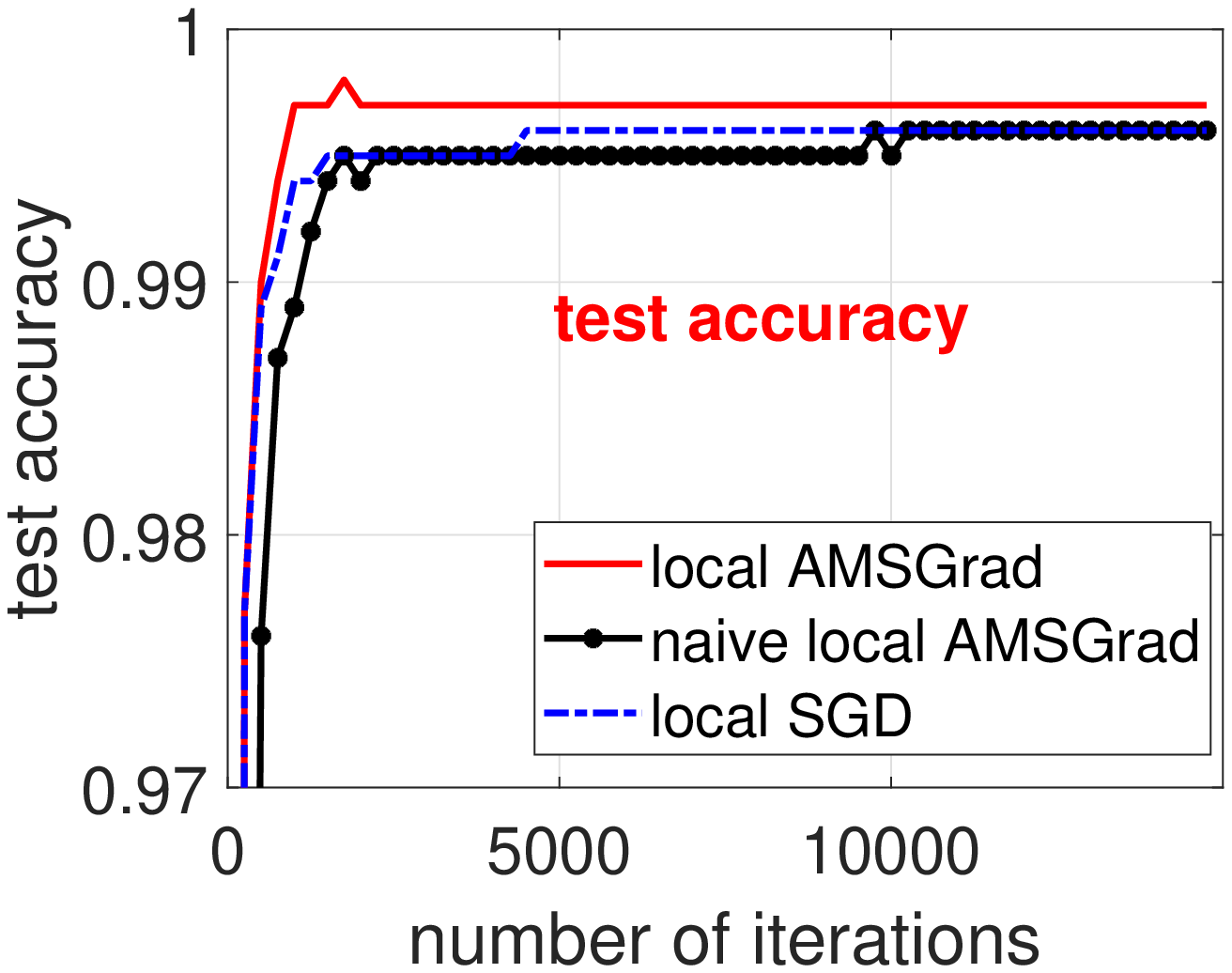}
}
\vspace{-0.1in}
	\caption{Gaussian mixture dataset: Performance comparison of three different algorithms.}
	\label{fig: compare_algorithms_s}
\end{figure}

We compare the performance of the algorithms with their best learning rate in Figure~\ref{fig: compare_algorithms_s}. It can be seen that local SGD and local AMSGrad perform very similarly. Naive local AMSGrad is worse than the other two algorithms by a small margin.

\newpage

\begin{figure}[h!]
\centering
\mbox{
	\includegraphics[width=2.2in]{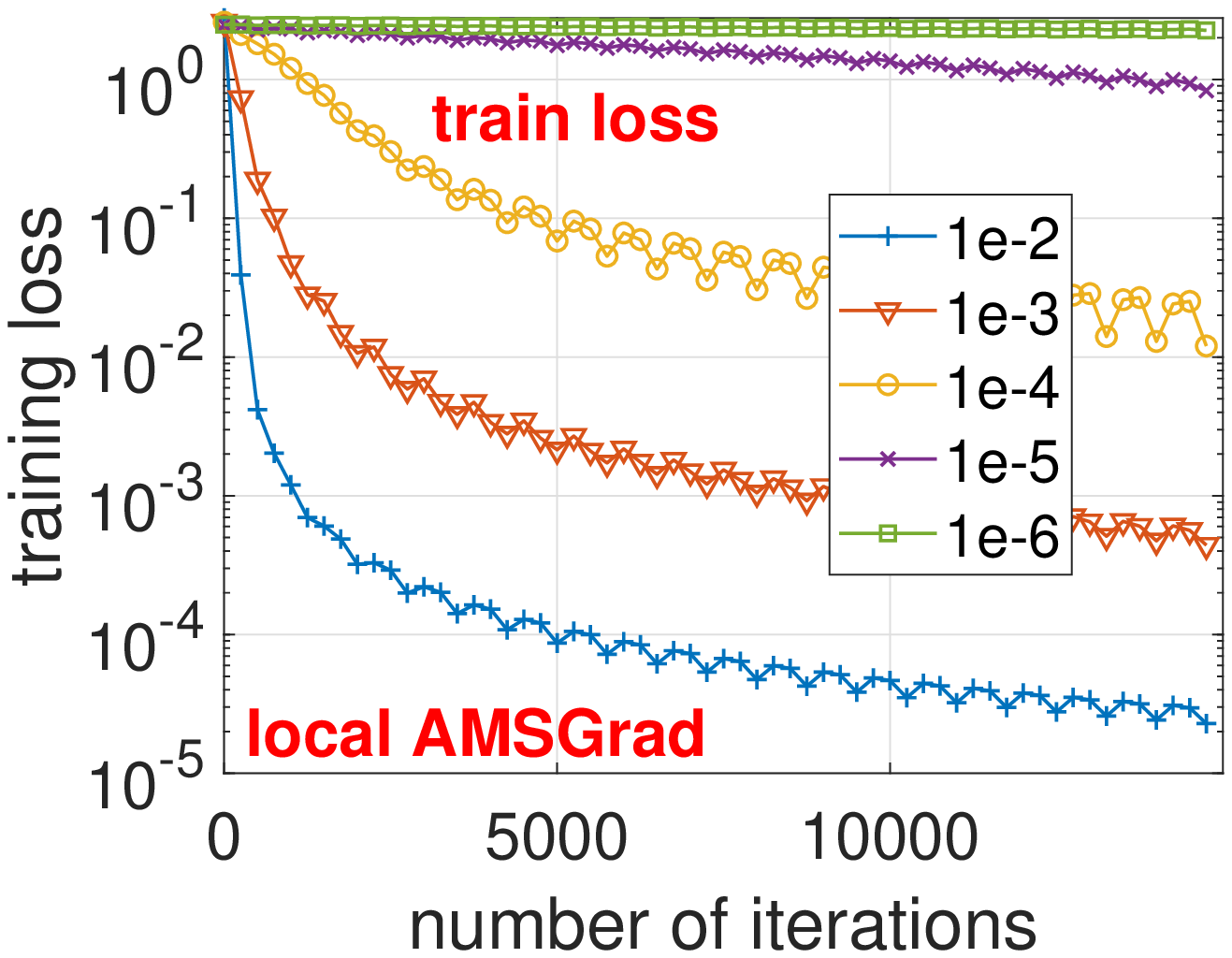}
	
	\includegraphics[width=2.2in]{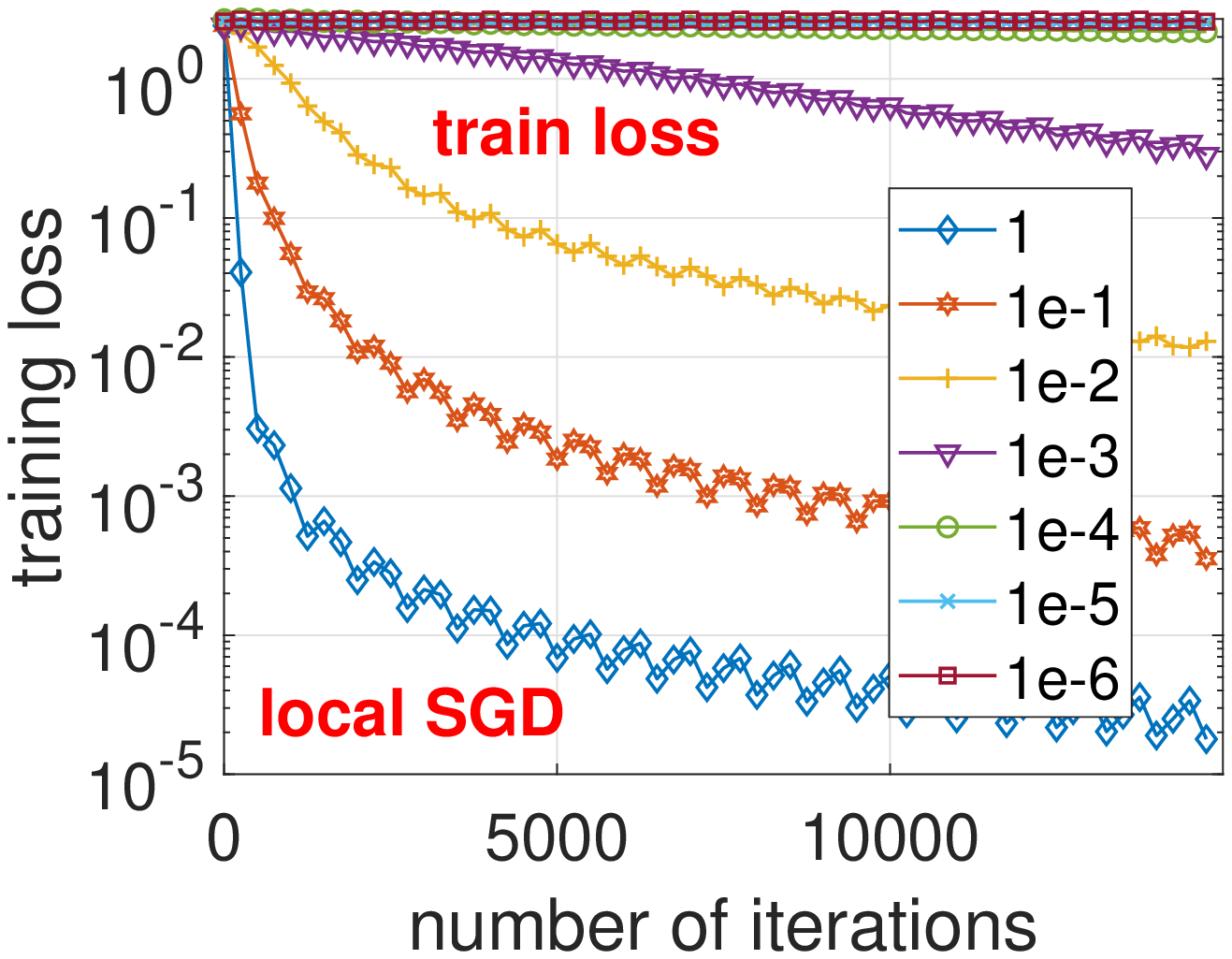}
	
	\includegraphics[width=2.2in]{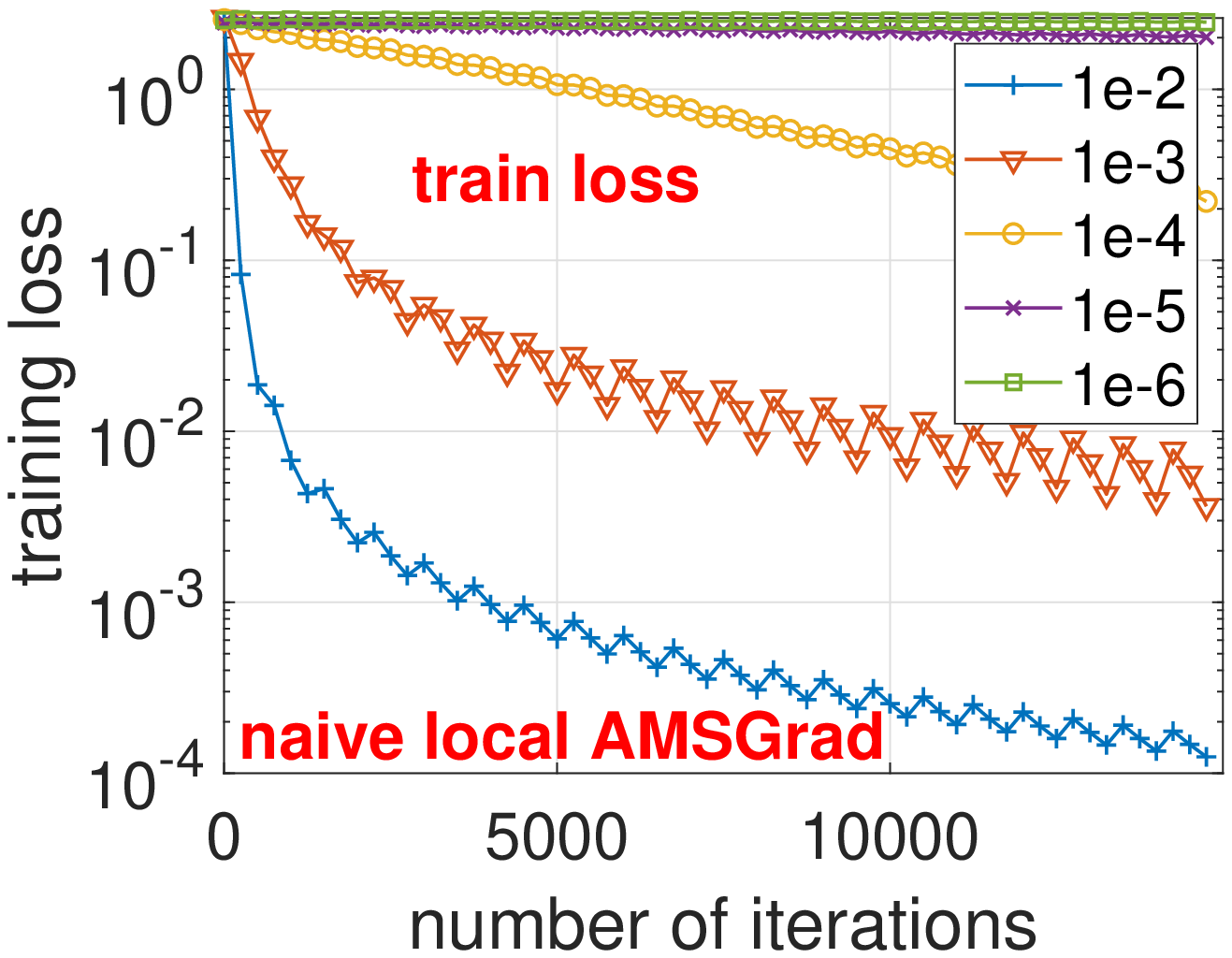}
}

\mbox{
	\includegraphics[width=2.2in]{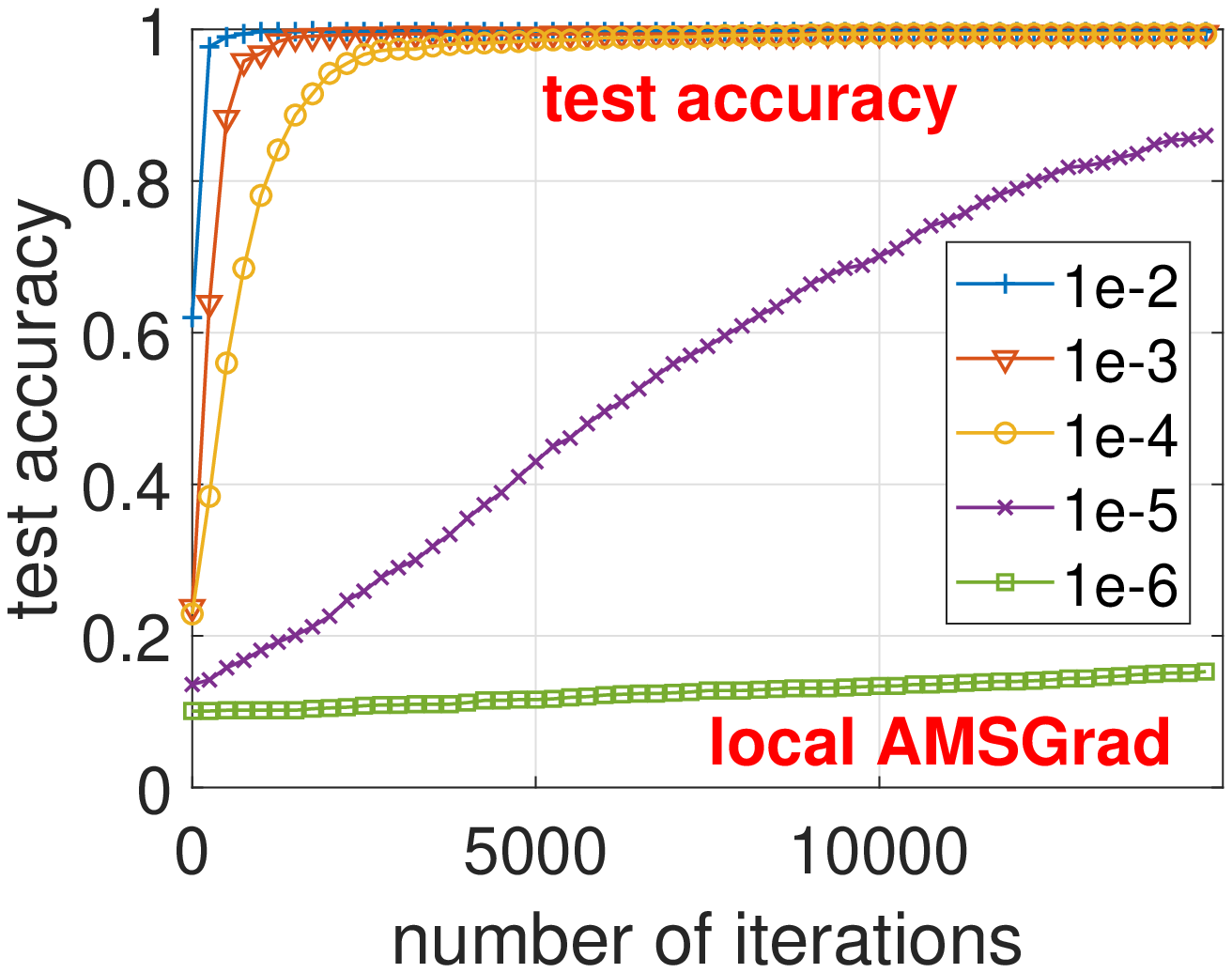}
	
	\includegraphics[width=2.2in]{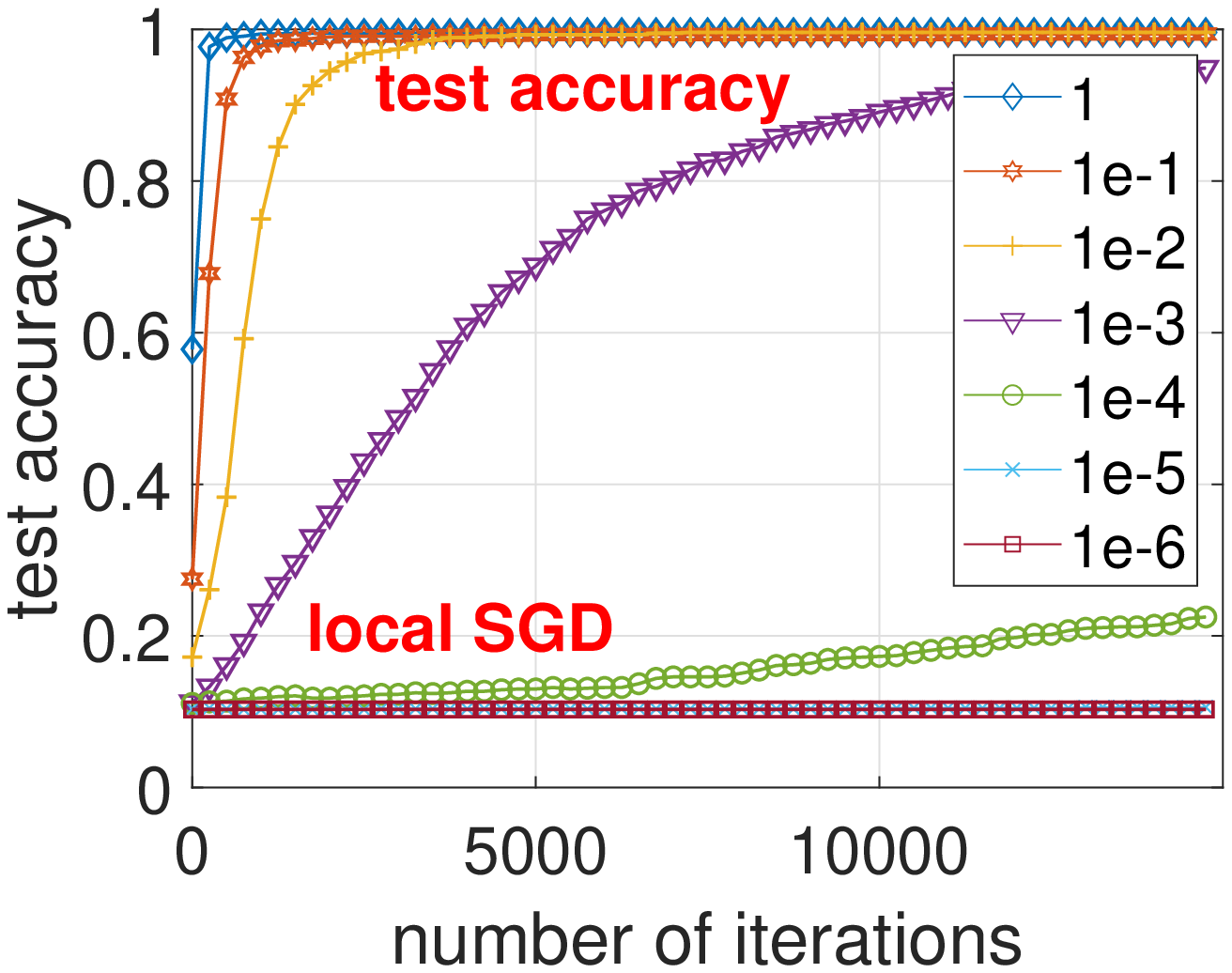}
	
	\includegraphics[width=2.2in]{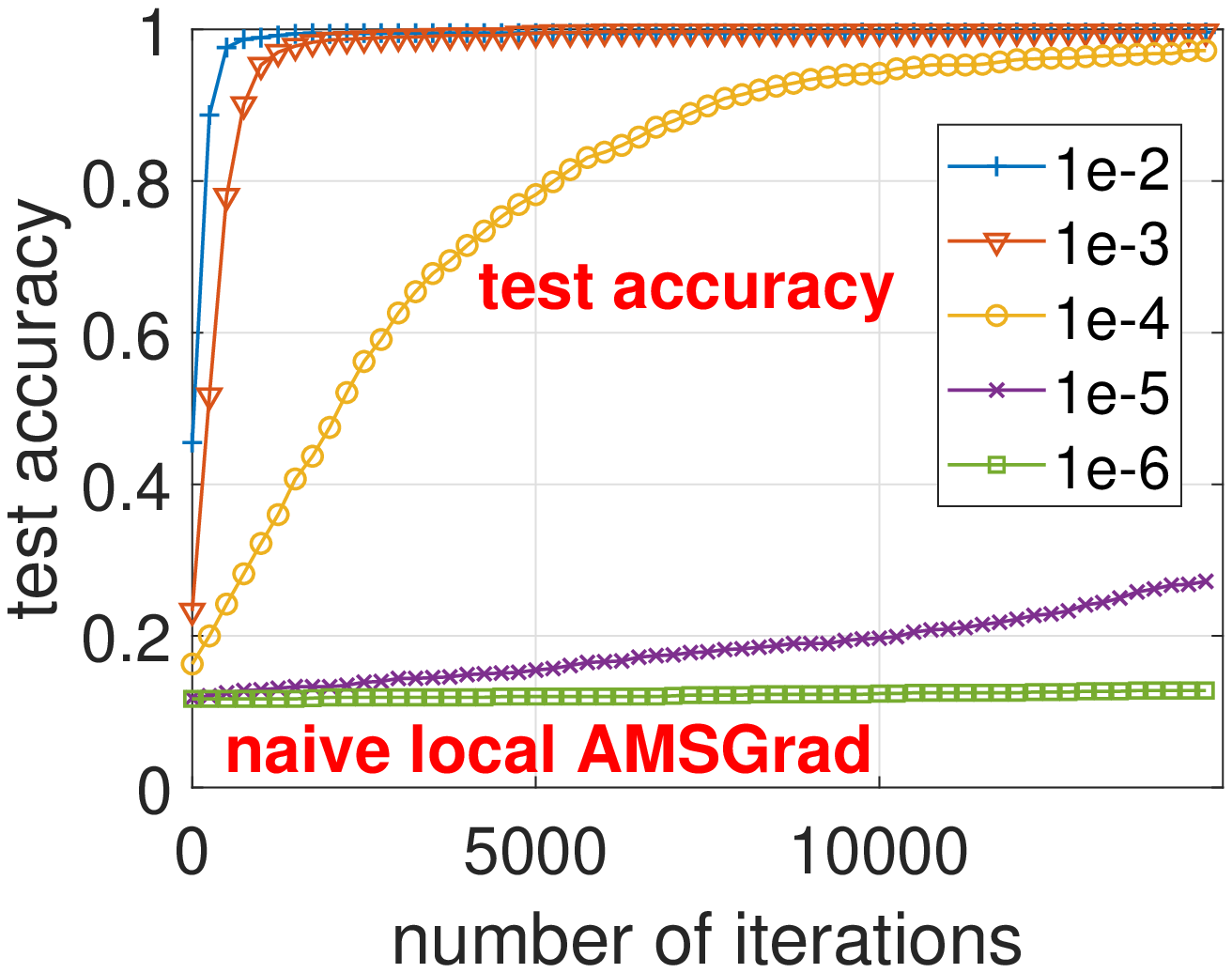}
}
		\vspace{-0.1in}
	\caption{Gaussian mixture dataset: Performance comparison of different stepsizes for three methods.}
	\label{fig: gauss_stepsize}
\end{figure}

The performance of different algorithms with different learning rate is shown in Figure~\ref{fig: gauss_stepsize}. It can be seen that all algorithms perform well with suitable learning rate. From these results, it seems that local AMSGrad do not have clear advantages over other algorithms. In particular, naive local AMSGrad performs not bad albeit it lacks convergence guarantee. We conjecture that this is due to that such a simple dataset cannot make the adaptive learning rate on different nodes differ significantly. In the next sets of experiments, we use more complicated real-world datasets to test the performance of different algorithms.

\newpage

\subsection{MNIST dataset (non-i.i.d. case)}

In this section, we compare the algorithms on training a convolutional neural network (CNN) on MNIST. Similar to last set of experiments, we perform learning rate search starting from 1e-6 until the algorithm diverges or deteriorates significantly. The average period $k$ is set to 10. We set $\epsilon$ to be 1e-4 for both Adam and AMSGrad. The data is distributed on 5 nodes and each node contains data from two classes, and there is no overlap on labels between different nodes. We expect such an allocation of data can creates a highly non-i.i.d. data distribution, leading to significantly different adaptive learning rates on different nodes. The neural network in the experiments consists of 3 convolution+pooling layers with ReLu activation, followed by a 10 nodes fully connected layer with softmax activation. The first convolution+pooling layer has 20 5x5 filters followed by 2x2 max pooling with stride 2. The second and third convolution+polling layer has 50 filters with other parameters being the same as the first layer.

\begin{figure}[h]
\centering
\mbox{
	\includegraphics[width=3in]{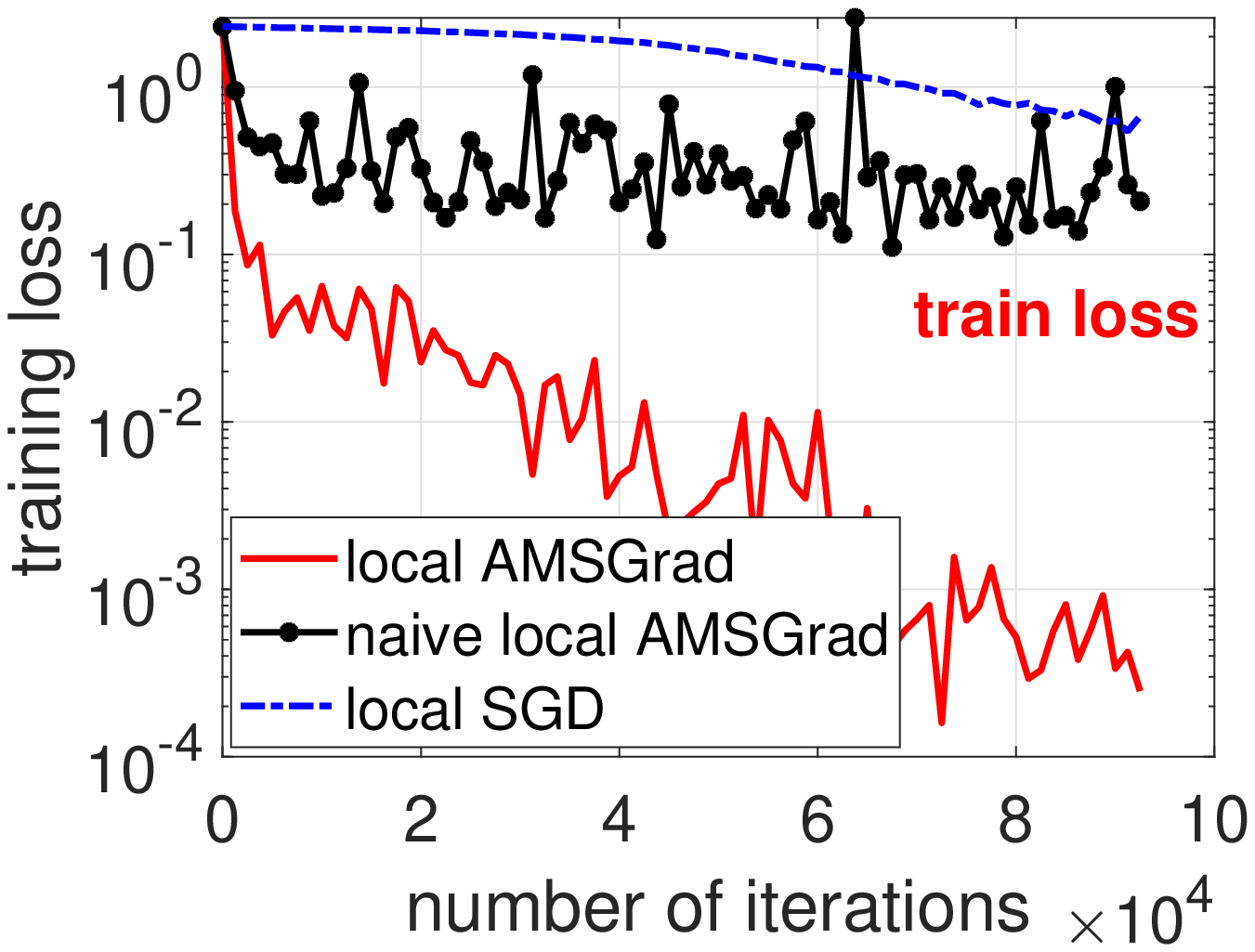}

		\includegraphics[width=3in]{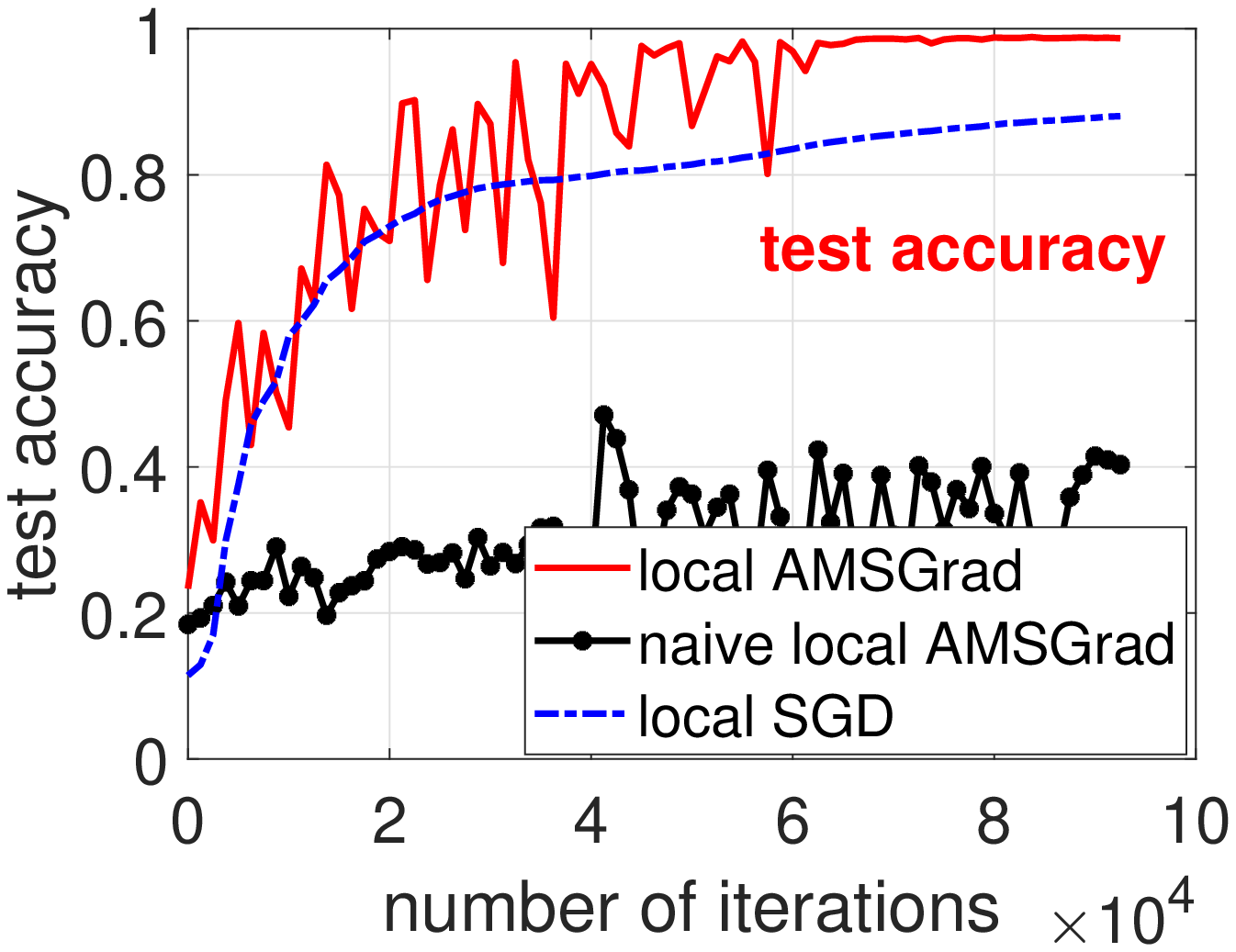}
}
	\caption{MNIST dataset: Performance comparison of three different algorithms.}
	\label{fig: compare_algorithms}
\end{figure}

Figure~\ref{fig: compare_algorithms} shows the training and testing performance of different algorithms. 
Specifically, 1e-3 is chosen for local AMSGrad and naive local AMSGrad while 1e-4 is chosen for local SGD. It can be seen that local AMSGrad outperforms local SGD by a large margin.

\begin{figure}[H]

\mbox{
	\includegraphics[width=2.2in]{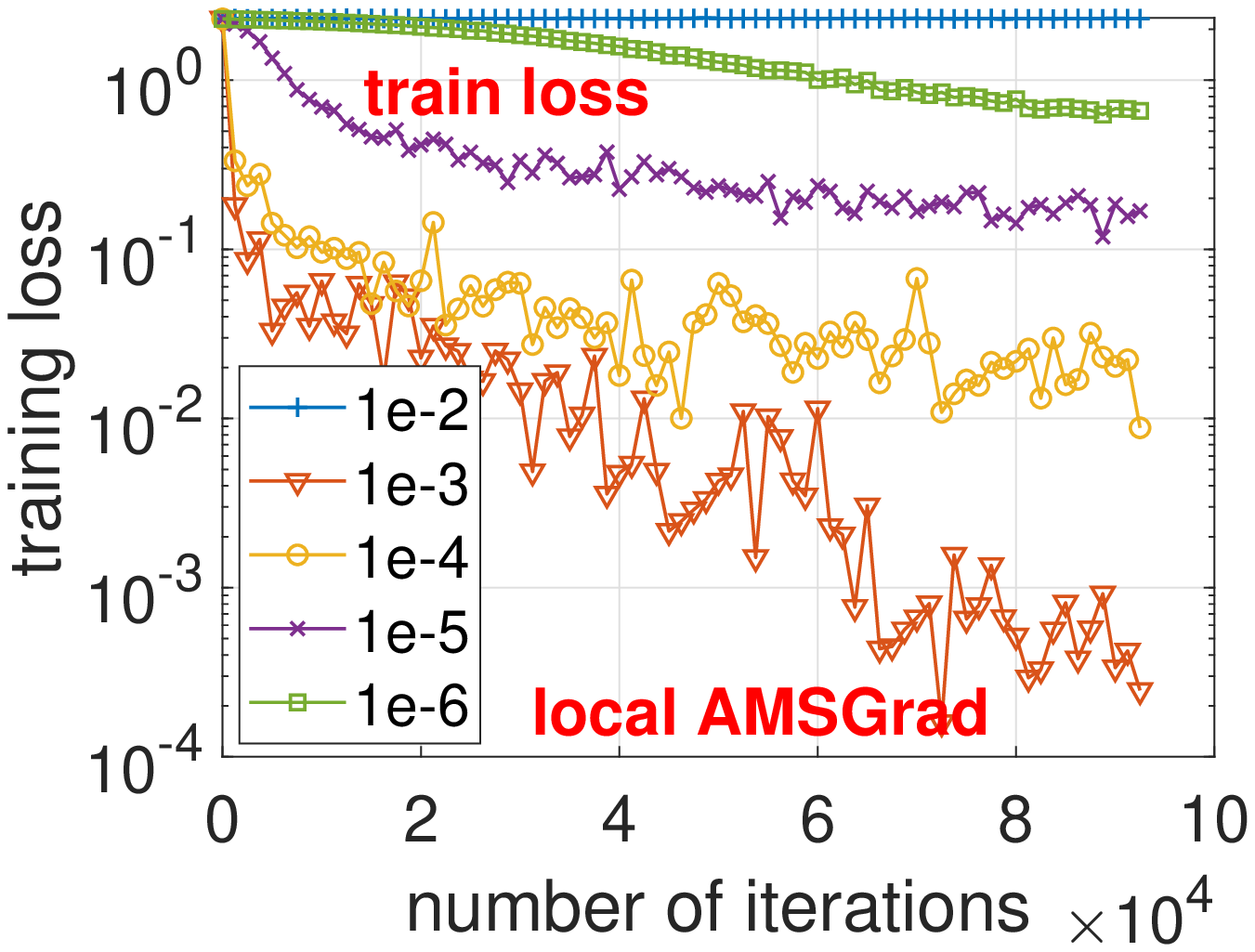}
	
	\includegraphics[width=2.2in]{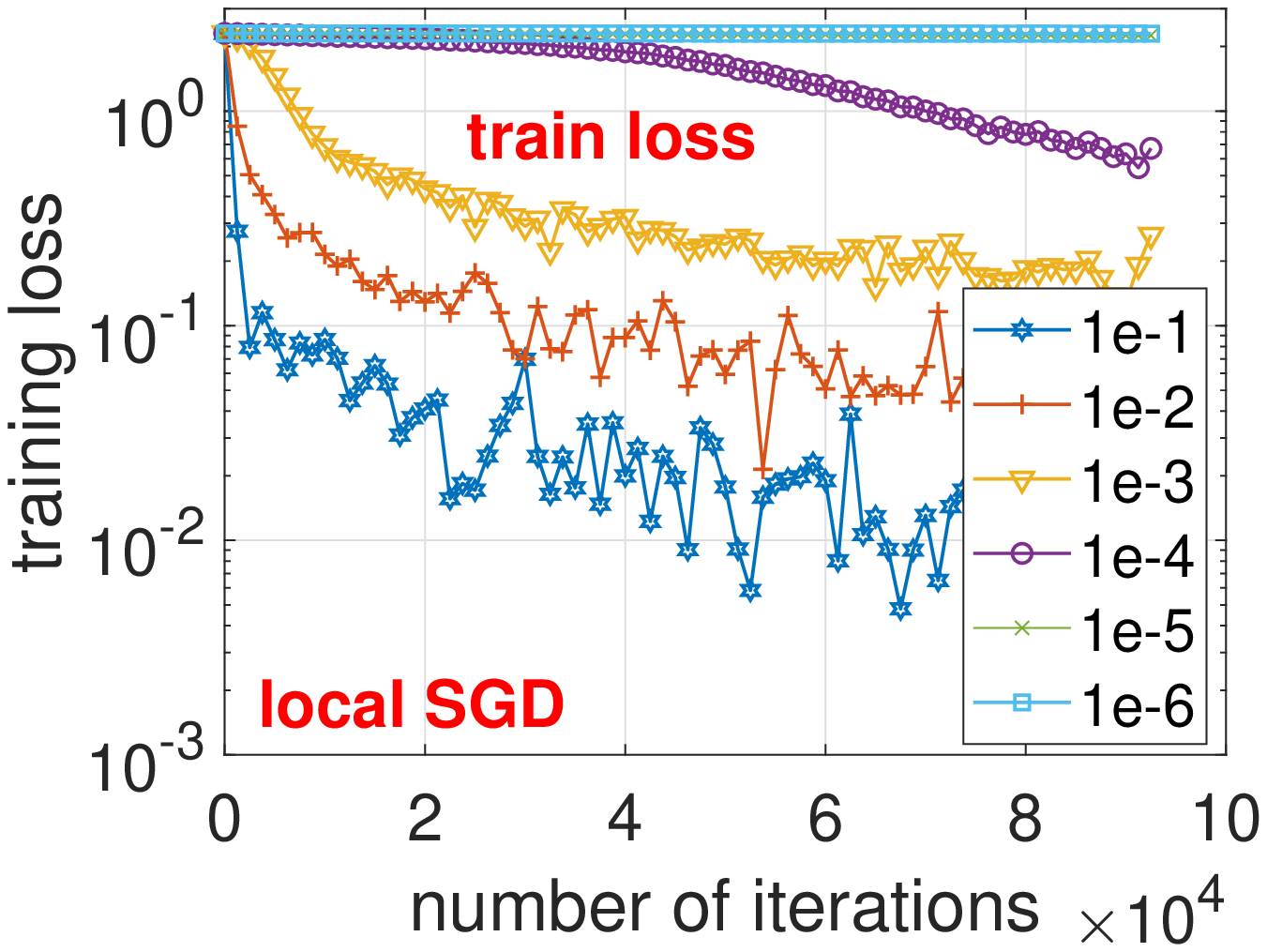}
	
	\includegraphics[width=2.2in]{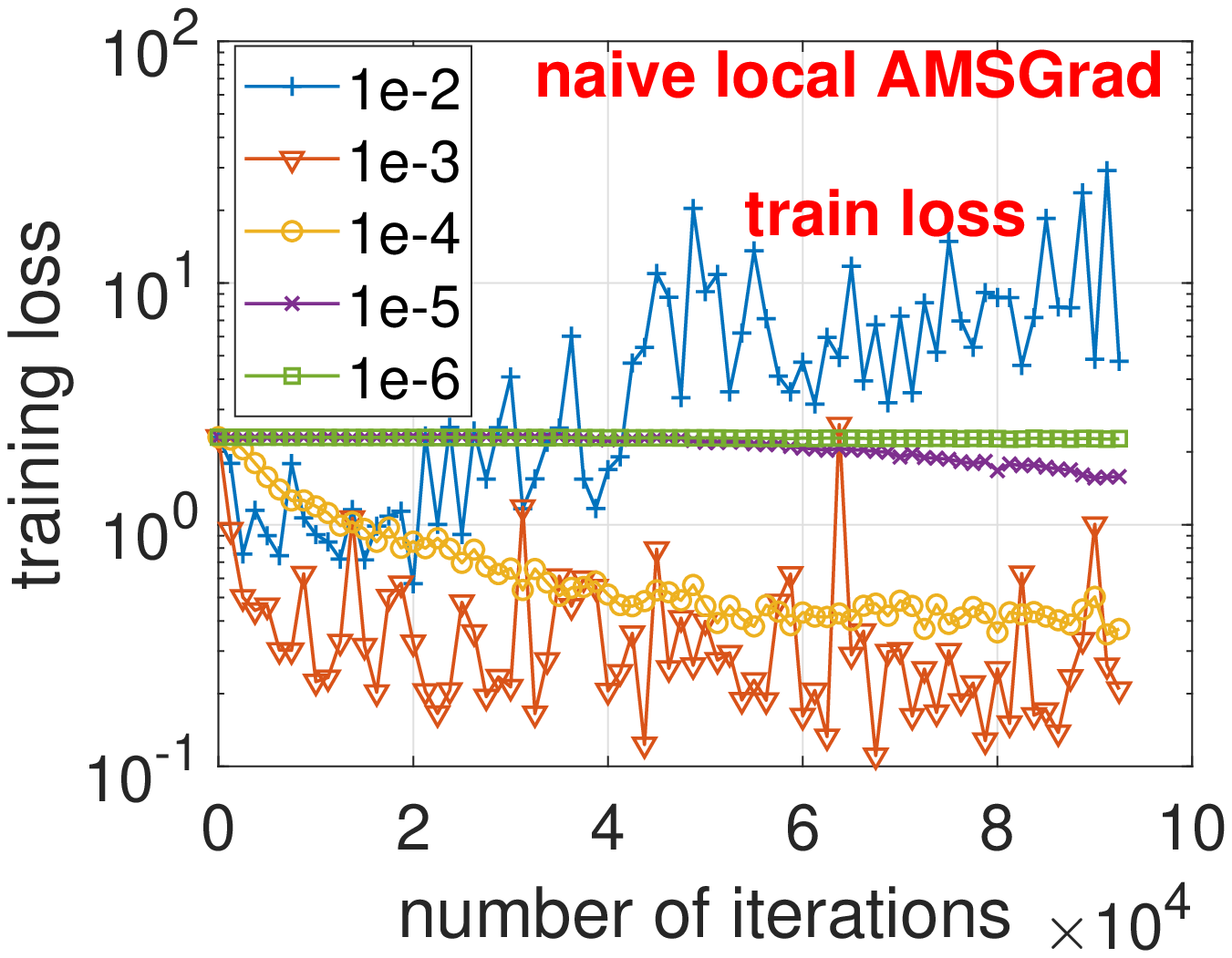}

}

\mbox{
		\includegraphics[width=2.2in]{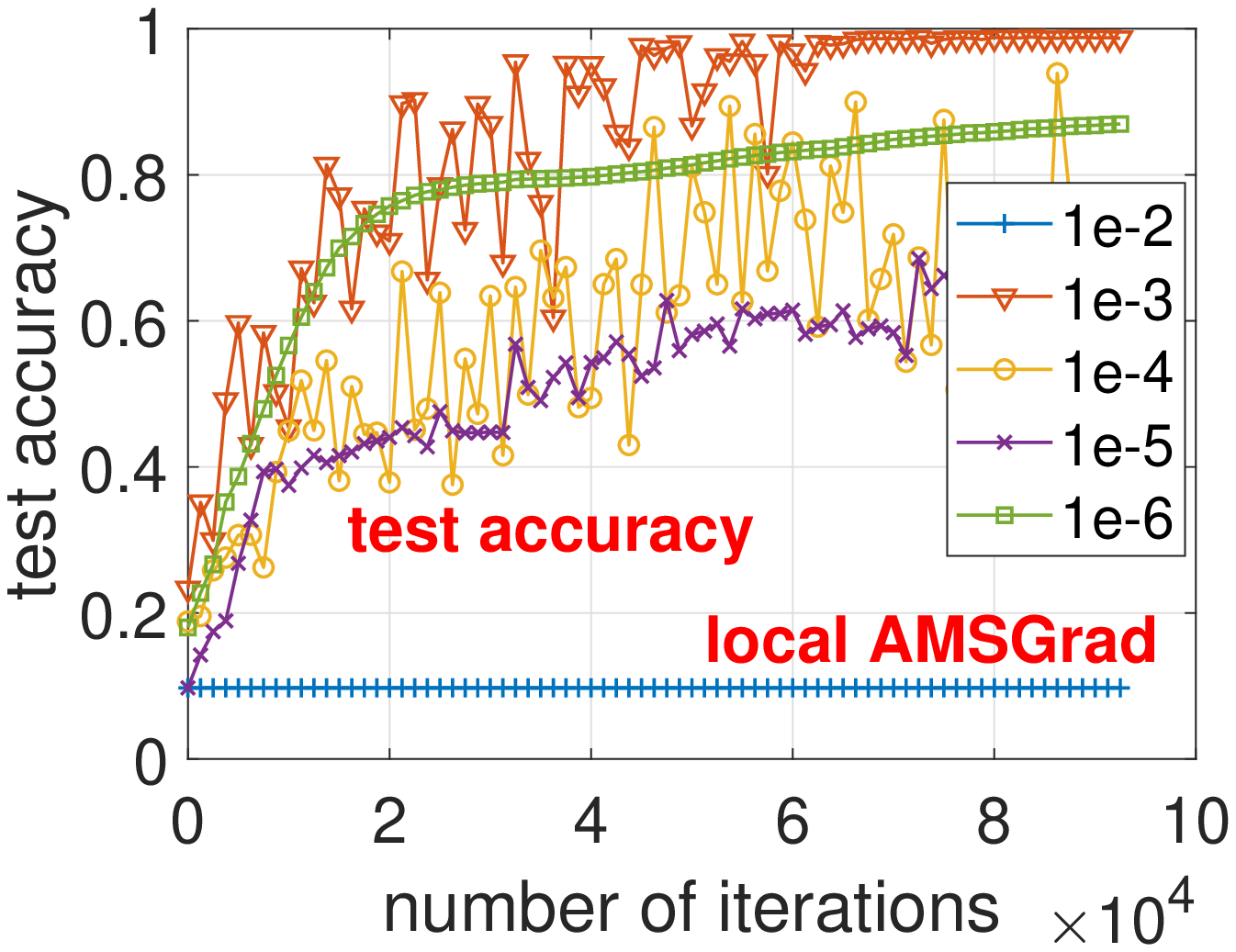}
		
		\includegraphics[width=2.2in]{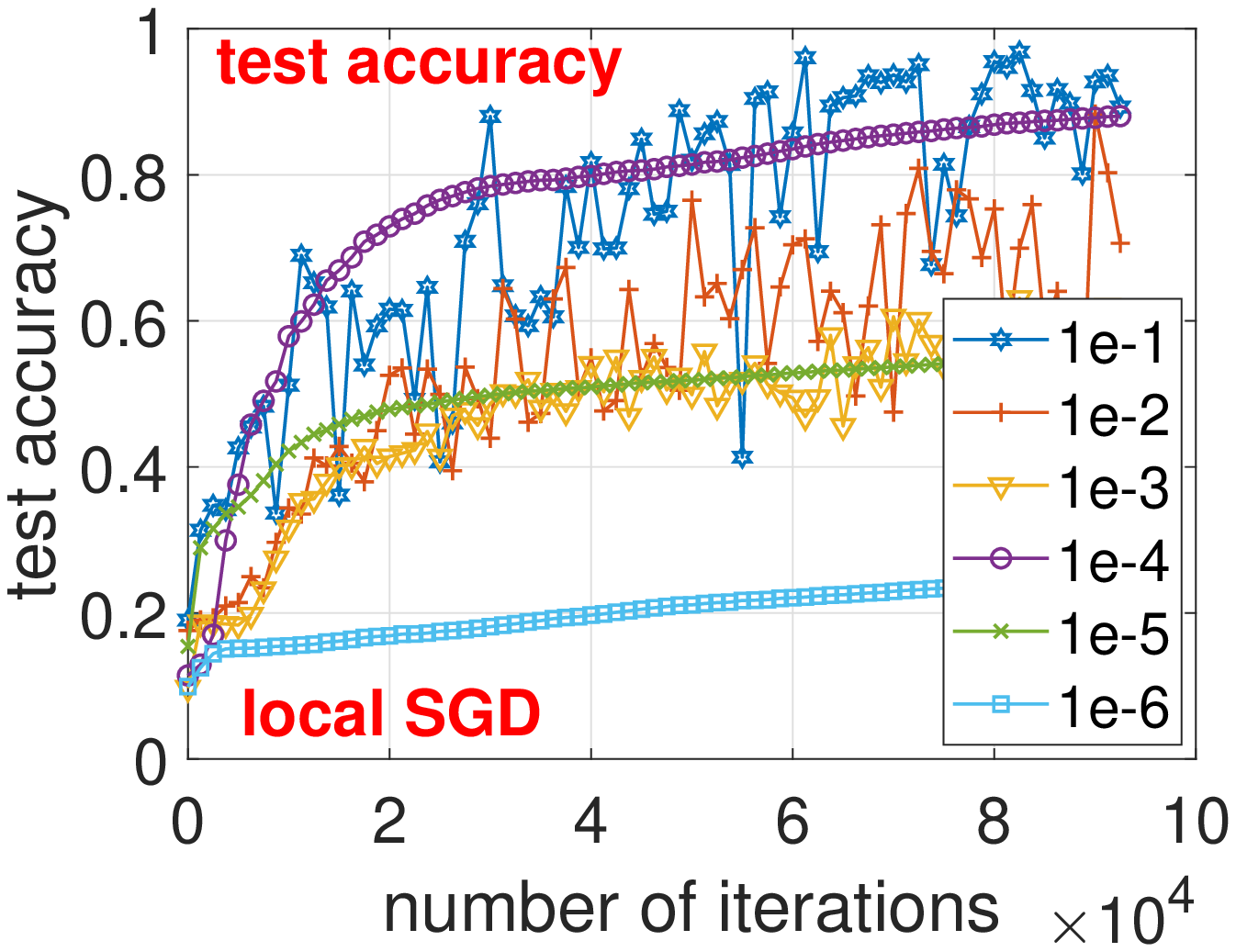}
		
		\includegraphics[width=2.2in]{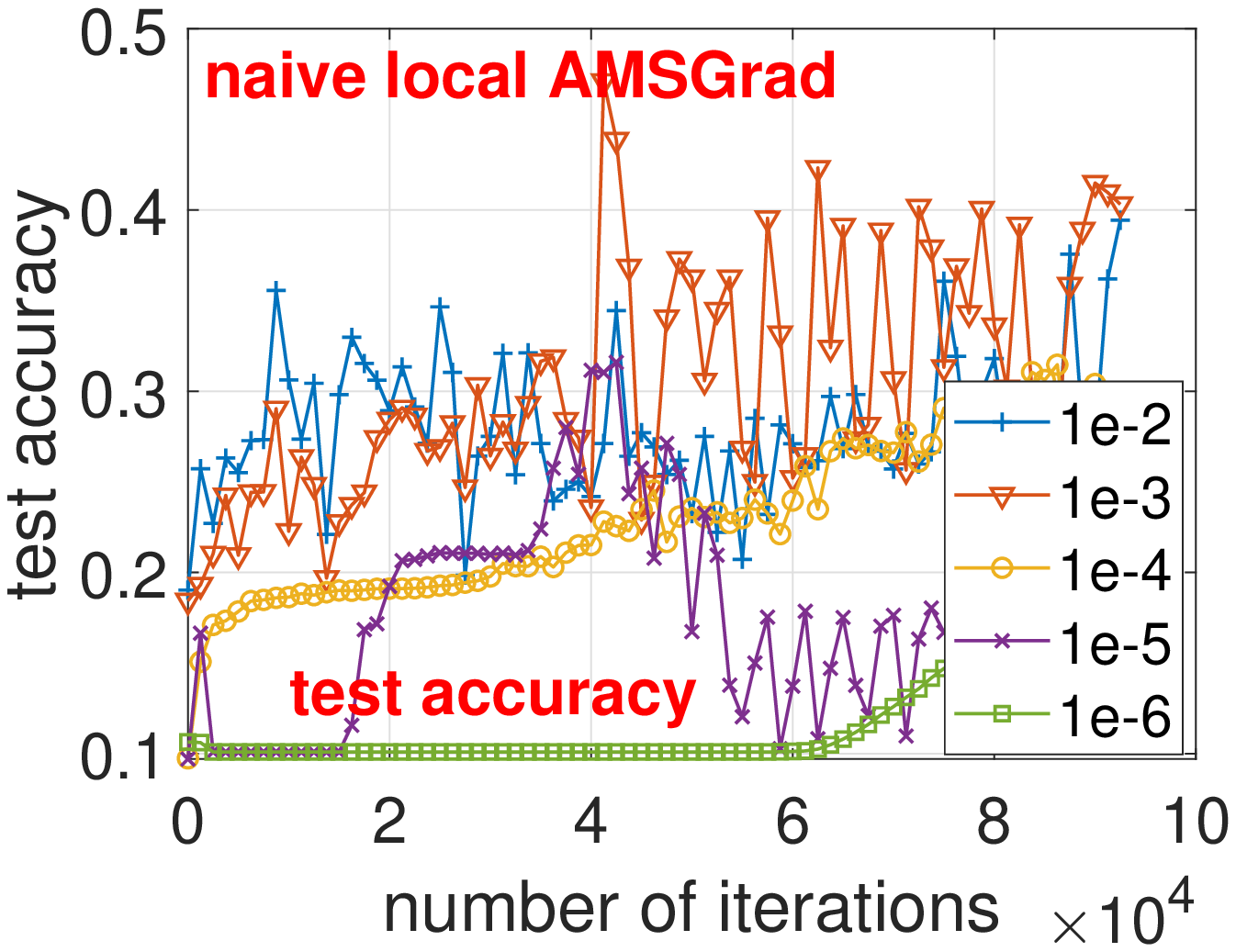}

}
		\vspace{-0.1in}
	\caption{MNIST dataset: Performance comparison of different stepsizes for three methods.}
	\label{fig: mnist_stepsize}
\end{figure}

The performance of algorithms with different learning rate is shown in Figure~\ref{fig: mnist_stepsize}. We observe that naive local AMSGrad has very poor performance with all learning rate choices, and local AMSGrad tend to perform better than local SGD on average. The slow convergence of local SGD is also observed in~\citet{mcmahan2016communication} when the data distribution is non-i.i.d. While the sampling on nodes in~\citet{mcmahan2016communication} may somehow reduce the influence of non-i.i.d. data, the convergence speed of local SGD is significantly impacted by the non-i.i.d. distribution in our experiment since we always use all nodes for parameter update.

\subsection{Letter recognition dataset (i.i.d. case)}

\begin{figure}[b!]
\centering
\mbox{ \includegraphics[width=2.8in]{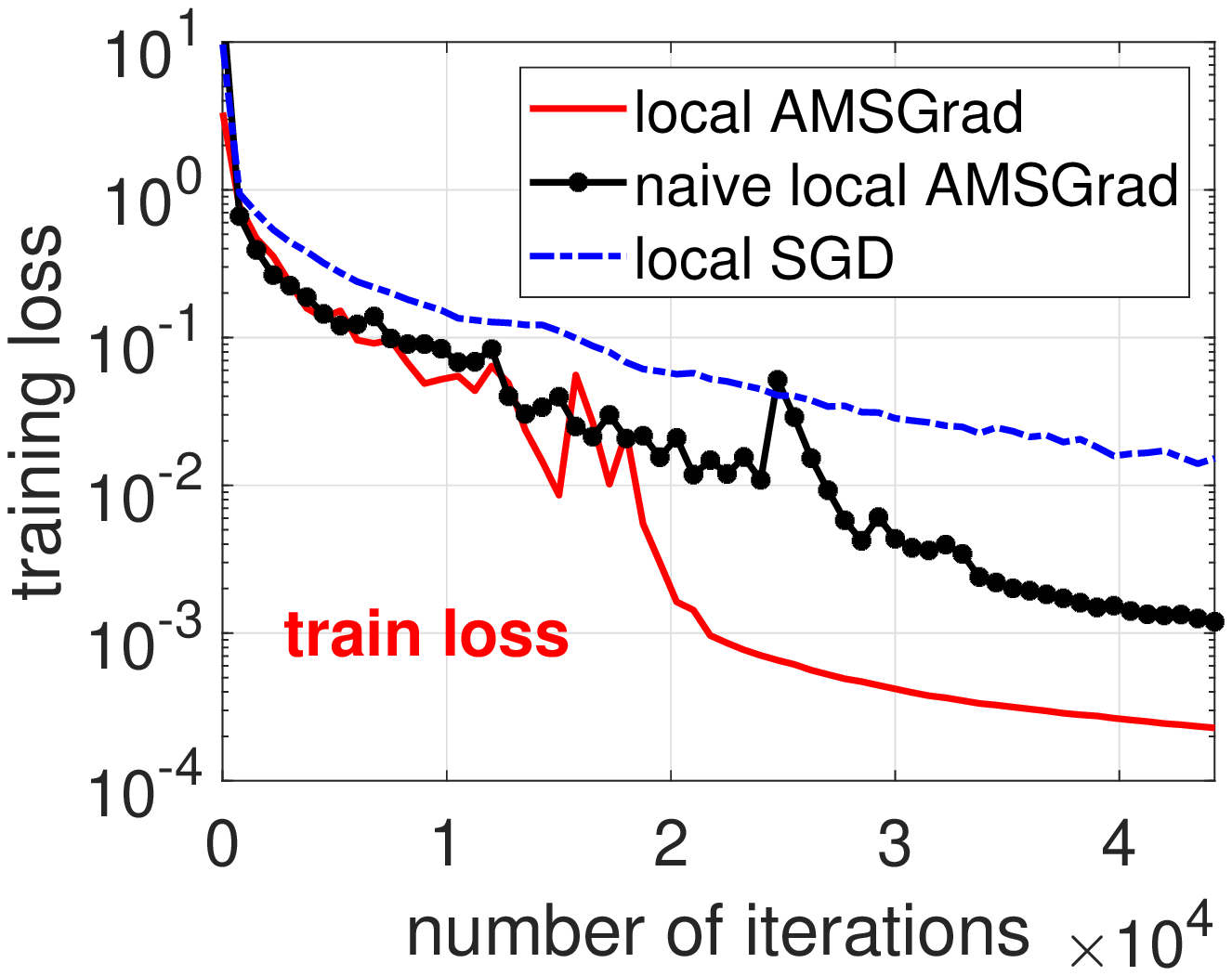}
 \hspace{0.15in}
		\includegraphics[width=2.8in]{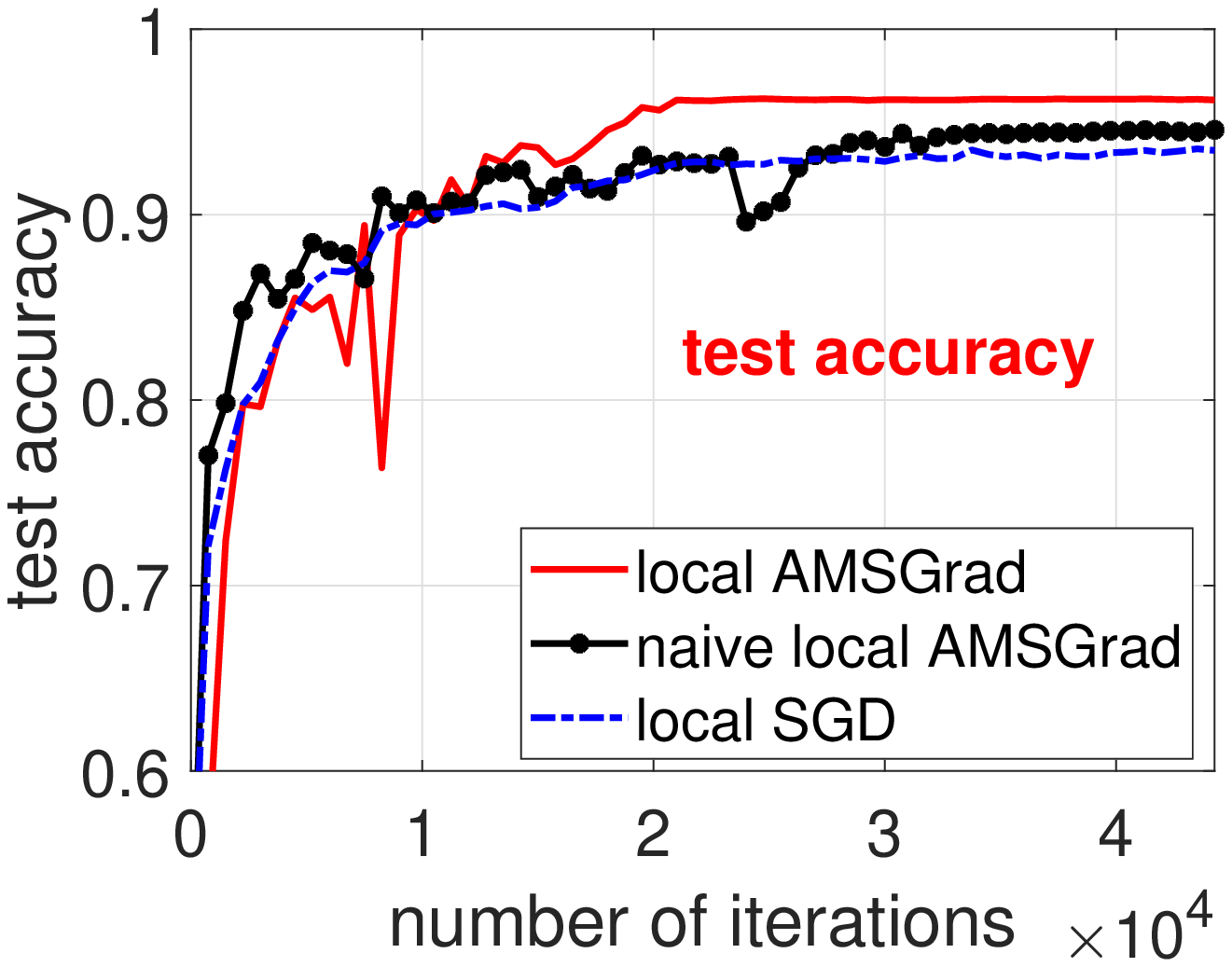}
}
\vspace{-0.1in}
	\caption{Letter recognition dataset: Performance comparison of three different algorithms.}
	\label{fig: compare_algorithms_l}
\end{figure}

\begin{figure}[t]
\centering

\mbox{
	\includegraphics[width=2.2in]{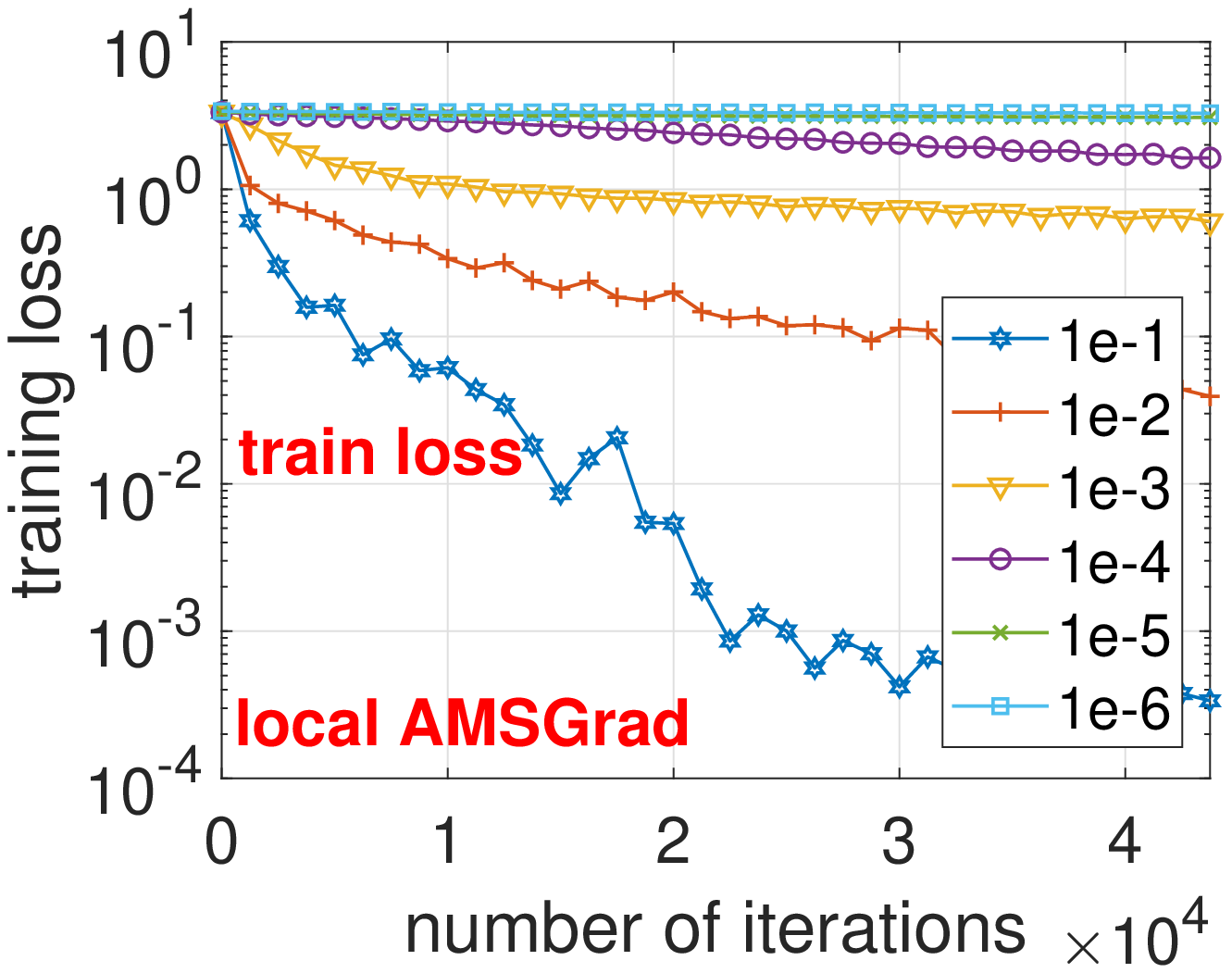}
	
	\includegraphics[width=2.2in]{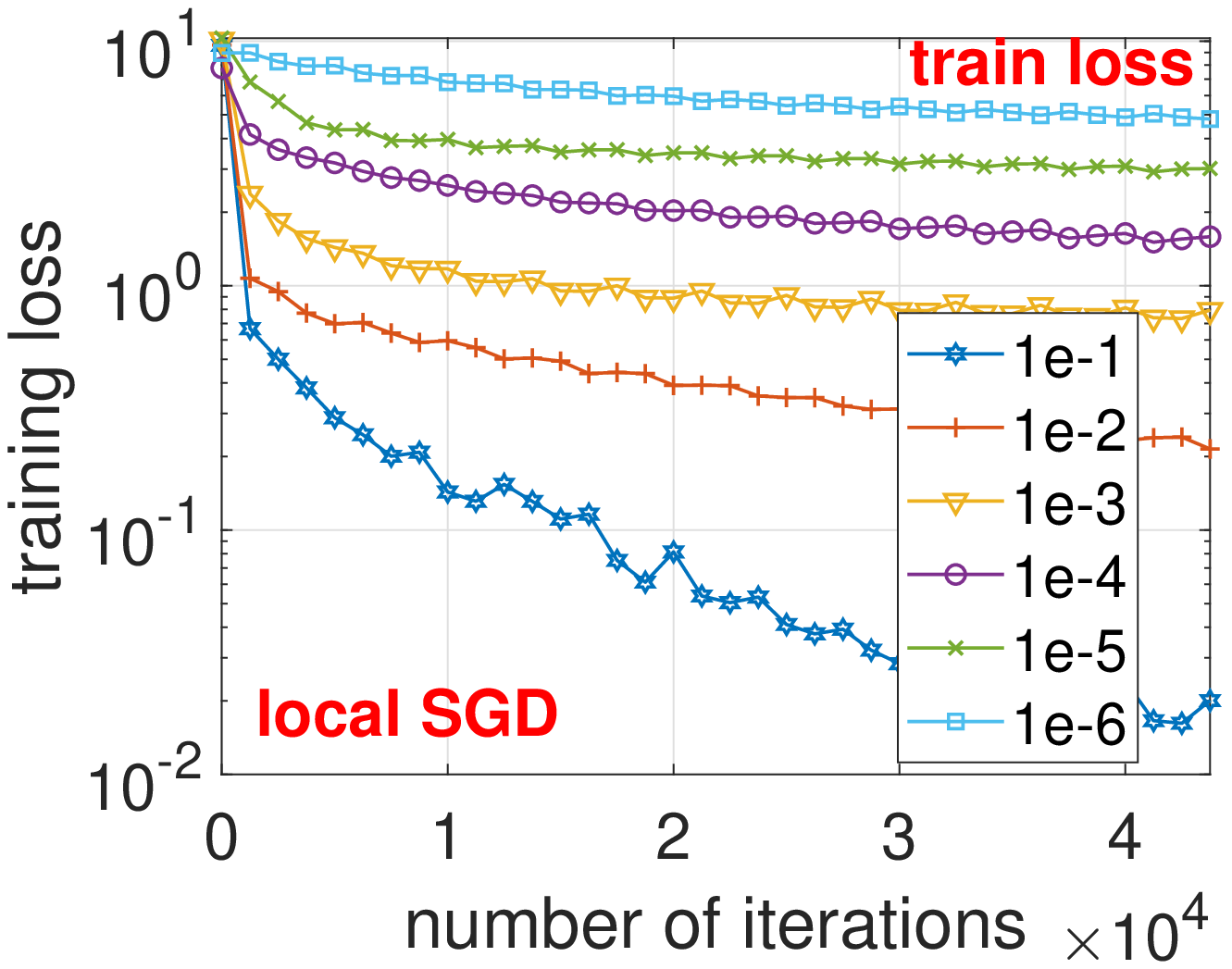}
	
	\includegraphics[width=2.2in]{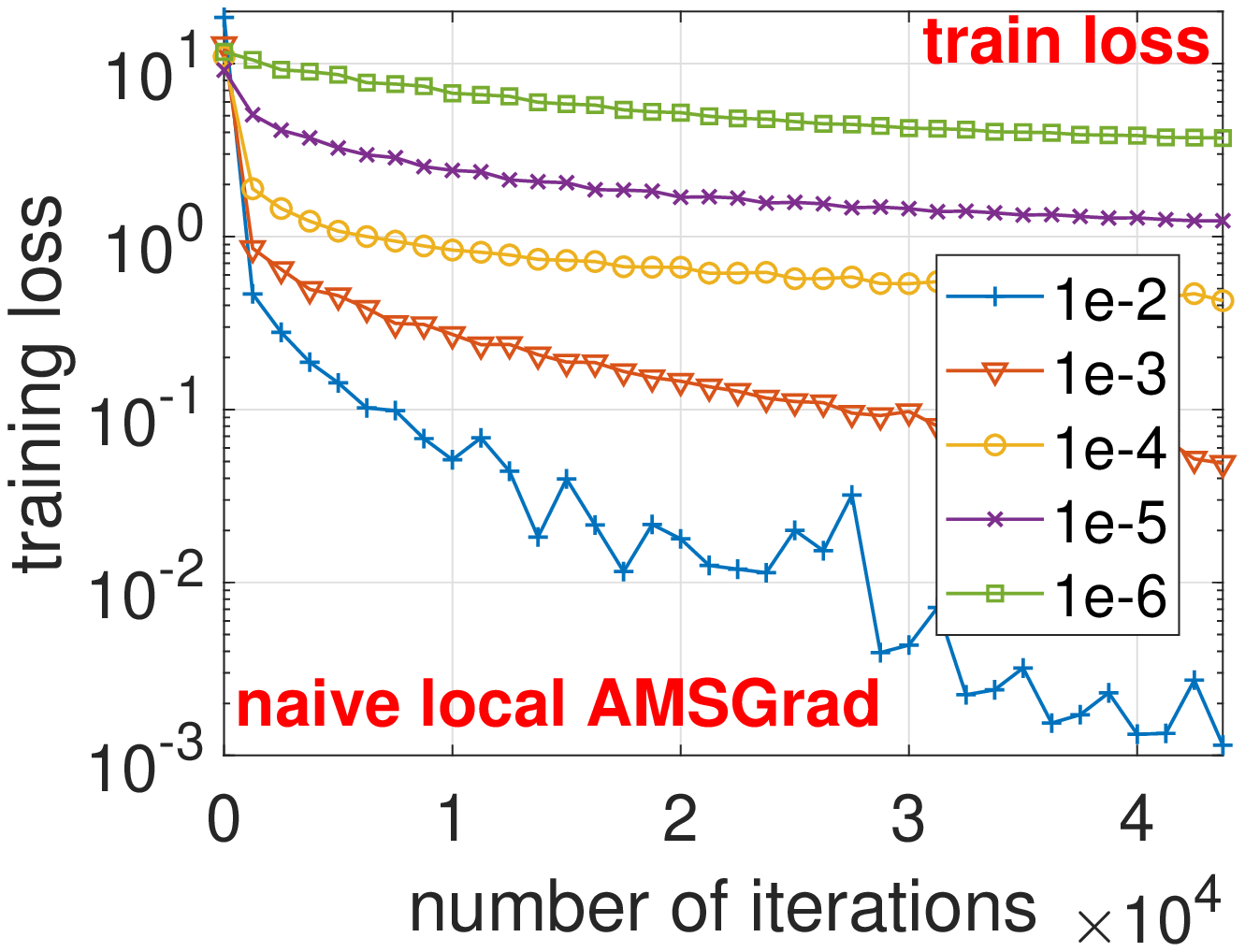}

}

\mbox{
		\includegraphics[width=2.2in]{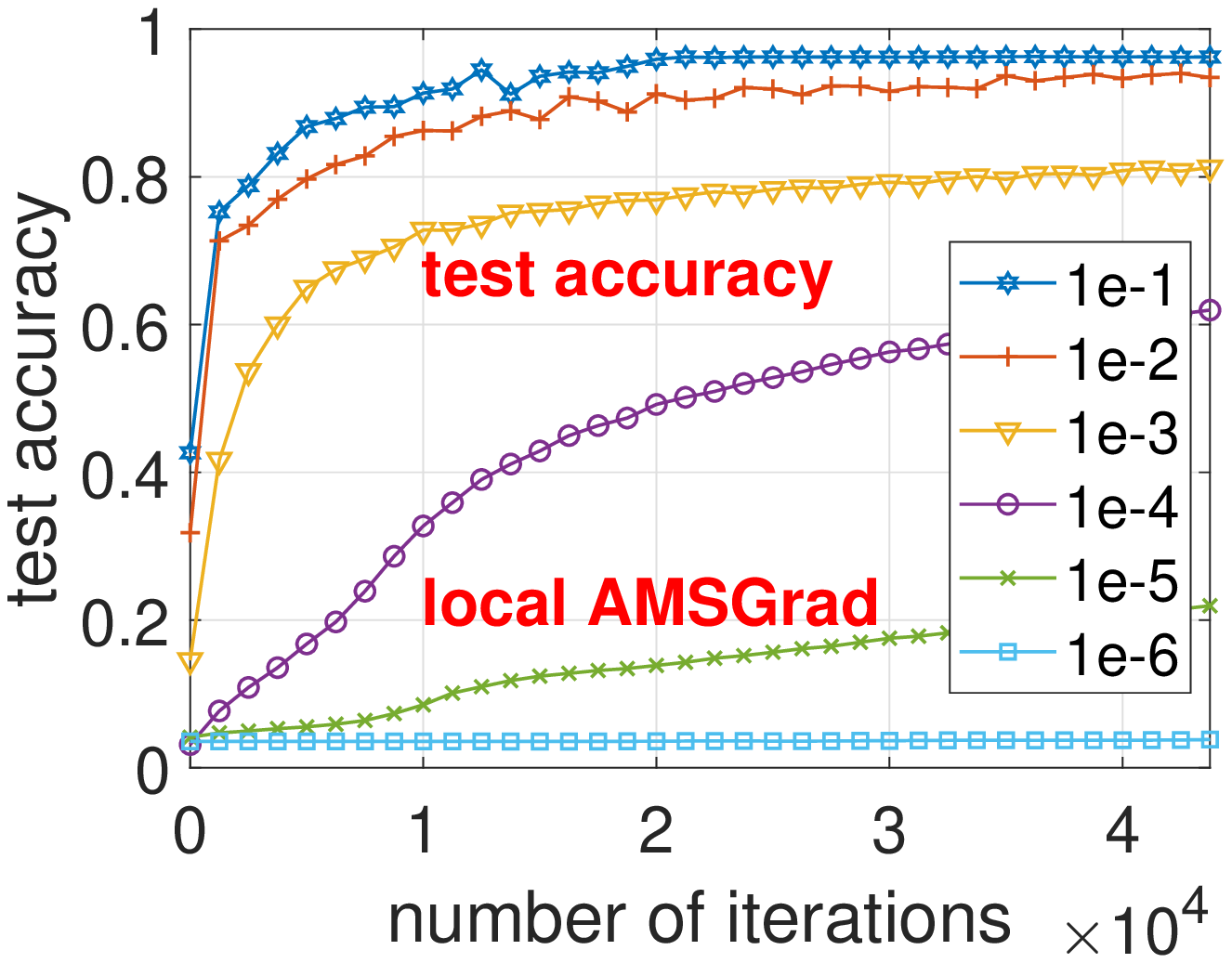}
	
		\includegraphics[width=2.2in]{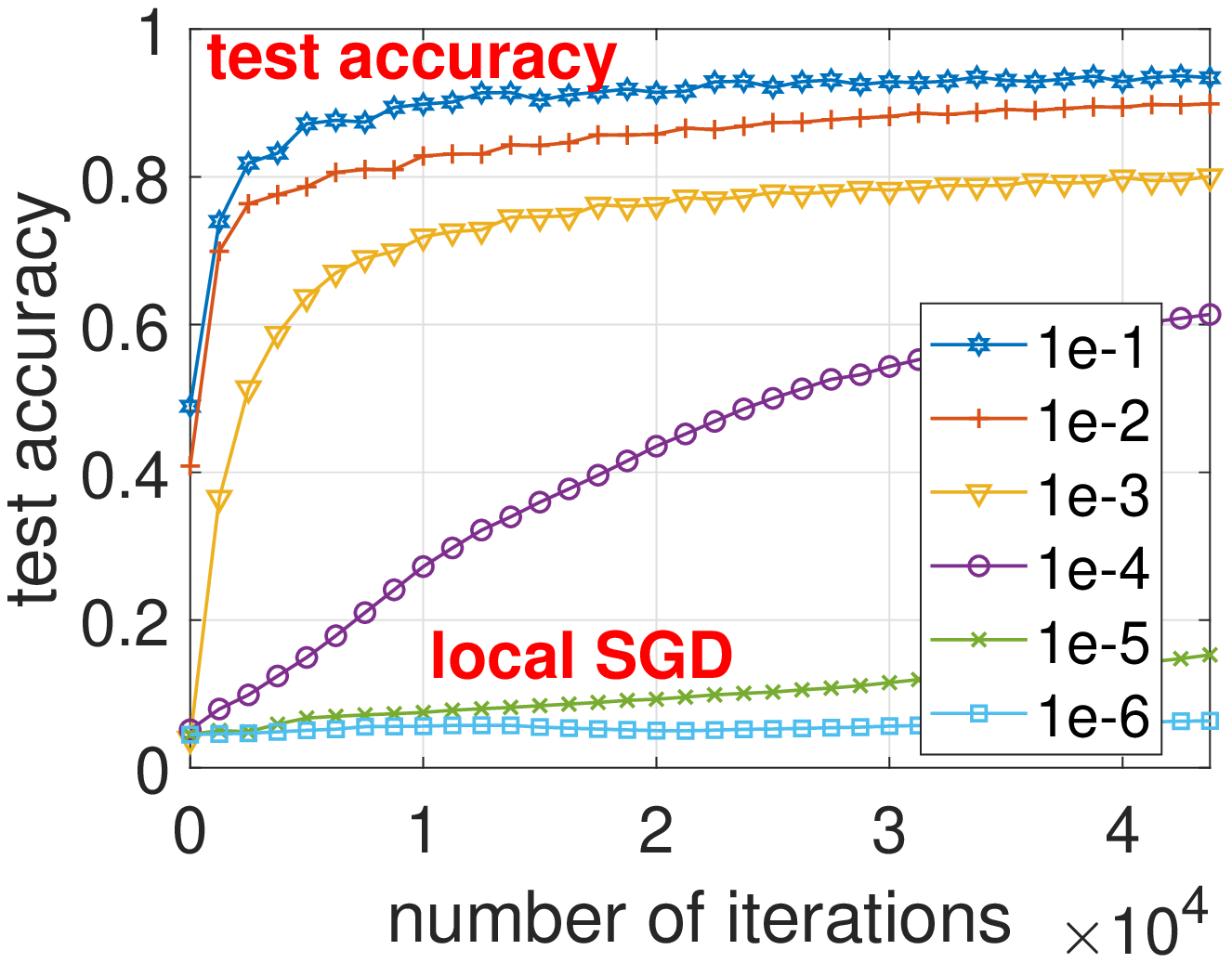}
	
		\includegraphics[width=2.2in]{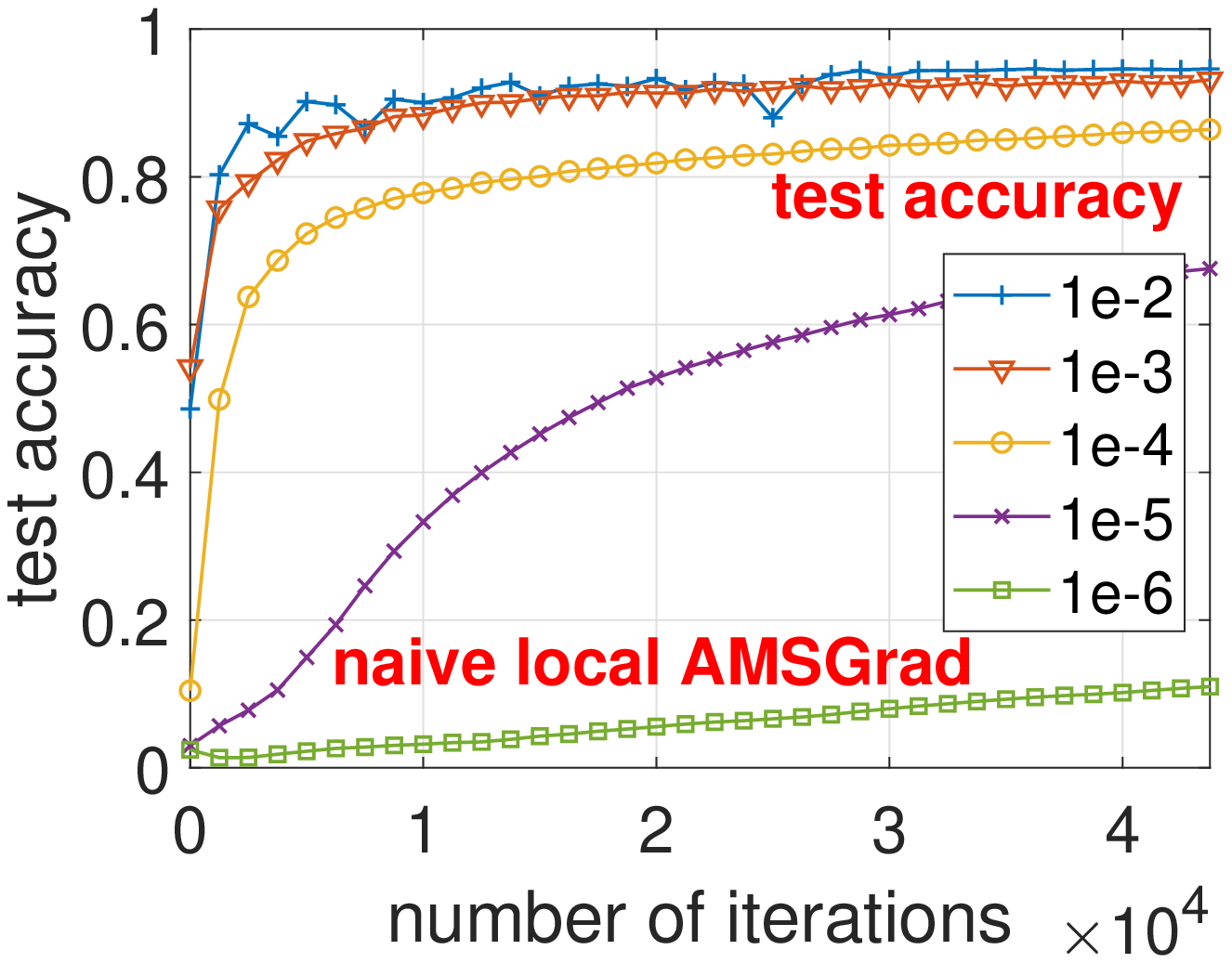}

}
		\vspace{-0.2in}
	\caption{Letter recognition dataset: Performance comparison of different stepsizes for three methods.}
	\label{fig: letter_stepsize}
\end{figure}

In this section, we use the letter recognition dataset~\cite{frey1991letter} to test the performance of different algorithms. We train a fully connected neural network with two hidden layers on the dataset. The first hidden layer has 300 nodes and the second layer has 200 nodes, both of which use ReLU as activation. The learning rate search strategy and other parameter settings are the same as the MNIST experiments. Different from the previous two sets of experiments, the data on the 5 workers are randomly assigned. This corresponds to an i.i.d. data distribution. Thus, all algorithms are expected to work well in this set of experiments. The performance comparison of algorithms with their best learning rate is provided in Figure~\ref{fig: compare_algorithms_l}. We can see that all algorithms achieve over 90\% accuracy and local AMSGrad again performs the best, with 2\% higher test accuracy than local SGD. In this case, naive local AMSGrad is also better than local SGD. Thus, when data distribution is i.i.d., it might be okay to use the naive version of local AMSGrad in practice, considering that it involves even less communication. The performance of algorithms with different learning rate is shown in Figure~\ref{fig: letter_stepsize}, which shows that all algorithms perform quite stable, while the two adaptive gradient methods still outperform local SGD in general.

\section{Conclusion}
In this paper, we study how to design adaptive gradient methods for federated learning by utilizing periodic model averaging. We first construct counter examples to illustrate how a naive combination of adaptive gradient methods and periodic model averaging can fail to converge. Then, by utilizing the insights from the study of non-convergence, we propose an adaptive gradient method \textit{local AMSGrad} in the setting of federated learning with proved convergence guarantee.\ \ \  Local AMSGrad enjoys  the sublinear~communication cost of periodic model averaging as well as the superb empirical performance of adaptive gradient methods. Experiments show that local AMSGrad can often significantly outperform both the local SGD and a naive design of local adaptive gradient methods, especially when the dada distribution on different nodes is non-i.i.d.

\clearpage
\newpage

\appendix
\noindent\textbf{\LARGE Appendix}\\

\section{Proof of Theorem~\ref{thm: loca_ams}}\label{app: proof_thm5}

To prove the convergence of local AMSGrad, we first define an auxiliary sequence of iterates
\begin{align}\label{eq: def_z}
 \overline{z}_t = \overline{x}_t + \frac{\beta_1}{1-\beta_1}(\overline x_t - \overline x_{t-1} ).
\end{align}
where $\overline x_t = \frac{1}{N} \sum_{i=1} ^N x_{t,i}$ and we define $x_0 \triangleq x_1$.

\vspace{0.1in}

We have the following property for the new sequence $\overline z_t$.
\begin{lemma}\label{lem: aux_seq}
For $\overline z_t $ defined in~\eqref{eq: def_z},
\begin{align}\label{eq: z_update}
    \overline z_{t+1} - \overline z_{t} = \alpha \frac{\beta_1}{1-\beta_1 } \left(\frac{1}{\sqrt{\hat v_{t-1}}} - \frac{1}{\sqrt{\hat v_{t}}}\right) \odot \overline m_{t-1} - \alpha \frac{\overline g_t}{\sqrt{\hat v_t}}.
\end{align}
where $\overline m_{t-1} = \frac{1}{N}  \sum_{i=1}^ N m_{t-1,i}$ and  $\overline g_{t} = \frac{1}{N} \sum_{i=1}^ N g_{t,i}$.
\end{lemma}
\noindent\textbf{Proof:}
We have
\begin{align*}
    \overline z_{t+1}-\overline z_t&=\overline x_{t+1}+\frac{\beta_1}{1-\beta_1}(\overline x_{t+1}-\overline x_t)-\overline x_t-\frac{\beta_1}{1-\beta_1}(\overline x_{t}-\overline x_{t-1})\\
    &=\frac{1}{1-\beta_1}(\overline x_{t+1}-\overline x_t)-\frac{\beta_1}{1-\beta_1}(\overline x_t-\overline x_{t-1})\\
    &=\frac{\alpha}{1-\beta_1}\frac{\overline m_t}{\sqrt{\hat v_t}}-\frac{\alpha\beta_1}{1-\beta_1}\frac{\overline m_{t-1}}{\sqrt{\hat v_{t-1}}}.
\end{align*}
By the updating rule of Algorithm~\ref{alg: local_ams}, we have $\overline m_t=\beta_1\overline m_{t-1}+(1-\beta_1)\overline g_{t}$. Thus,
\begin{align*}
    \overline z_{t+1}-\overline z_t&=\frac{\alpha\beta_1}{1-\beta_1}\left( \frac{1}{\sqrt{\hat v_t}}-\frac{1}{\sqrt{\hat v_{t-1}}} \right)\odot \overline m_{t-1}+\alpha\frac{\overline g_t}{\sqrt{\hat v_t}},
\end{align*}
which completes the proof. \hfill $\square$

\vspace{0.05in}

In what follows, we use the auxiliary sequence $\bar z_t$ in Lemma \ref{lem: aux_seq} to prove convergence of the algorithm.

By Lipschitz continuity, we have
\begin{align}
    f(\overline  z_{t+1}) \leq     f(\overline  z_{t}) + \langle \nabla f(\overline z_t ), \overline z_{t+1} - \overline z_t \rangle + \frac{L}{2 } \|\overline z_{t+1} - \overline z_t \|^2, \nonumber
\end{align}
and thus
\begin{align}\label{eq: lipschitz_exp}
      -\mathbb E [ \langle \nabla f(\overline z_t ), \overline z_{t+1} - \overline z_t \rangle]
      \leq  & \mathbb E[   f(\overline  z_{t})] -  \mathbb E [f(\overline  z_{t+1})] + \frac{L}{2 }\mathbb E[ \|\overline z_{t+1} - \overline z_t \|^2].
\end{align}
where the expectation is taken over all the randomness of stochastic gradients until iteration $t$.

\newpage

It remains to upper bound the second-order term on RHS of~\eqref{eq: lipschitz_exp} and characterize the effective descent in the first order term (LHS of~\eqref{eq: lipschitz_exp}). We first characterize the effective descent.

By~\eqref{eq: z_update}, we can write the first-order term as
\begin{align}\label{eq: first_order}
     \langle \nabla f(\overline z_t ), \overline z_{t+1} - \overline z_t \rangle
    =    &  \alpha \langle  \nabla f(z_t), \frac{\beta_1}{1-\beta_1 } \left(\frac{1}{\sqrt{\hat v_{t-1}}} - \frac{1}{\sqrt{\hat v_{t}}}\right) \odot \overline m_{t-1} \nonumber \\
     - &\alpha \langle \nabla f(z_t), \frac{\overline g_t}{\sqrt{\hat v_t}} \rangle.
\end{align}
Since $\hat v_{t}$ is independent of $\overline g_t$ and $\mathbb E[g_{t,i}] = \nabla f_i(x_{t,i})$, taking expectation on both sides of~\eqref{eq: first_order} yields
\begin{align}
     \mathbb E \left[ \langle \nabla f(\overline z_t ), \overline z_{t+1} - \overline z_t \rangle \right]  
    =    &  \alpha \mathbb E \left [ \langle  \nabla f(z_t), \frac{\beta_1}{1-\beta_1 } \left(\frac{1}{\sqrt{\hat v_{t-1}}} - \frac{1}{\sqrt{\hat v_{t}}}\right) \odot \overline m_{t-1}\right ]  \nonumber \\
     - &\alpha \mathbb E \left[ \langle \nabla f(z_t), \frac{ \overline {\nabla f} (x_t)}{\sqrt{\hat v_t}} \rangle  \right], \nonumber
\end{align}
where $\overline {\nabla f} (x_t) = \frac{1}{N}\sum_{i=1}^N \nabla f_i (x_{t,i})$.

Using the fact $\langle a,b \rangle  = \frac{1}{2} (\|a-b\|^2 - \|a\|^2 - \|b\|^2 )$, we have
\begin{align}\label{eq: bias_split}
    \langle \nabla f(\overline z_t), \frac{ \overline {\nabla f} (x_t)}{\sqrt{\hat v_t}} \rangle = & \frac{1}{2}\left \| \frac{\nabla f(\overline z_t)}{\hat v_t^{1/4}}\right\|^2 + \frac{1}{2}\left\| \frac{\overline {\nabla f} (x_t)}{\hat v_t^{1/4}} \right\|^2 - \left \| \frac{\nabla f(\overline z_t) - \overline {\nabla f} (x_t)}{\hat v_t^{1/4}} \right\|^2,
\end{align}
where the first two quantities on RHS of~\eqref{eq: bias_split} will contribute to the descent of objective in a single optimization step, while the last term is the possible ascent introduced by the bias on the stochastic gradients. The bound of the last term is given by Lemma~\ref{lem: bias_overall_bound}.
\begin{lemma}\label{lem: bias_overall_bound}
For Algorithm~\ref{alg: local_ams}, we have
\begin{align}\label{eq: bias_overall_bound}
   \left \| \frac{\nabla f(\overline z_t) - \overline {\nabla f} (x_t)}{\hat v_t^{1/4}} \right\|^2
    \leq & \frac{L}{\min_j (\hat v_t^{1/2})_j} \left(2 \frac{\beta_1^2}{(1-\beta_1)^2} +8(k-1)^2  \right) \alpha^2  d \frac{G^2}{\epsilon}.
\end{align}
\end{lemma}
\noindent\textbf{Proof:} First, we have
\begin{align} \notag
    & \left \| \frac{\nabla f(\overline z_t) - \overline {\nabla f} (x_t)}{\hat v_t^{1/4}} \right\|^2\\\notag
    =&   \left \| \frac{\frac{1}{N} \sum_{i=1}^N ( \nabla f_i(\overline z_t) -  {\nabla f_i} (x_{t,i}))}{\hat v_t^{1/4}} \right\|^2   \\\notag
    \leq & \frac{1}{N} \sum_{i=1} ^ N   \left \| \frac{\nabla f_i(\overline z_t) -  {\nabla f_i} (x_{t,i})}{\hat v_t^{1/4}} \right\|^2  \\\label{eq: bias_bound}
    \leq & \frac{1}{N} \sum_{i=1} ^ N   2 \left(\left \| \frac{\nabla f_i(\overline z_t) -  \nabla f_i(\overline x_{t})}{\hat v_t^{1/4}} \right\|^2  + \left \| \frac{\nabla f_i(\overline x_t) -  {\nabla f_i} (x_{t,i})}{\hat v_t^{1/4}} \right\|^2 \right),
\end{align}
where the last inequality is due to Cauchy-Schwartz.

Using Lipschitz property (Assumption A1) of $\nabla f_i$, we can further bound the differences of gradients on RHS of~\eqref{eq: bias_bound} by
{
\begin{align}\label{eq: z_x_lip}
   \frac{2}{N} \sum_{i=1} ^ N    \left \| \frac{\nabla f_i(\overline z_t) - \nabla f_i(\overline x_{t})}{\hat v_t^{1/4}} \right\|^2
    \leq \frac{2}{N} \sum_{i=1} ^ N \frac{\left \| {\nabla f_i(\overline z_t) - \nabla f_i(\overline x_{t})}{} \right\|^2  }{\min_{j} (\hat v_t^{1/2})_j}
    \leq  \frac{2}{N} \sum_{i=1} ^ N \frac{L \|\overline z_t -   \overline x_{t}\|^2  }{\min_{j} (\hat v_t^{1/2})_j}.
\end{align}
}%
Similarly, we have
\begin{align}\label{eq: x_x_avg_diff}
       & \frac{2}{N} \sum_{i=1} ^ N \left \| \frac{\nabla f_i(\overline x_t) -  {\nabla f_i} (x_{t,i})}{\hat v_t^{1/4}} \right\|^2
    \leq \frac{2}{N} \sum_{i=1} ^ N  \frac{L\| \overline x_{t} - x_{t,i}\|^2 }{\min_{j} (\hat v_t^{1/2})_j}.
\end{align}

It remains to bound $\|\overline z_t - \overline x_t\|^2$ and $\|\overline x_t - x_{t,i}\|^2$ using the update rule of $x$ and $z$.
For the difference between $\overline z_t$ and $\overline x_t$, we have
\begin{align} \label{eq: z_x_bound}
\sum_{i=1} ^ N  \left \| {\overline z_t -   \overline x_{t}}{} \right\|^2  = &\frac{\beta_1^2}{(1-\beta_1)^2} \sum_{i=1} ^ N   \left \| {\overline x_t - \overline x_{t-1}}{} \right\|^2 \nonumber \\
=&  \frac{\beta_1^2}{(1-\beta_1)^2}   \alpha^2  N\left\| \frac{\overline m_{t-1}}{ \sqrt{\hat v_{t-1}}}\right\|^2 \nonumber \\
\leq& \frac{\beta_1^2}{(1-\beta_1)^2} \alpha^2 Nd  \frac{G^2}{\epsilon},
\end{align}
where $\overline m_t = \frac{1}{N}\sum_{i=1}^N m_{t,i}$. For the second term containing the consensus error $\overline x_t - x_{t,i}$, we have
{
\begin{align}\label{eq: consensus_bound}
    \sum_{i=1} ^ N  \left \| {\overline x_t - x_{t,i}}{} \right\|^2
    \leq & 4 N(k-1)^2 \alpha^2 d \frac{G^2}{\epsilon}
\end{align}
}%
by Lemma~\ref{lem: z_x_diff}, which is presented next.

\begin{lemma}\label{lem: z_x_diff}
For iterates produced by Algorithm~\ref{alg: local_ams}, we have
\begin{align}\notag
     \left \| {\overline x_t - x_{t,i}}{} \right\|^2
    \leq & 4 (k-1)^2 \alpha^2 d \frac{G^2}{\epsilon}, \vspace{0.2in}  \forall i \in [N].
\end{align}
\end{lemma}

\noindent\textbf{Proof}: Let $\floor{t}_k$ be the largest multiple of $k$ that is less than $t$. By the updating rule of Algorithm \ref{alg: local_ams}, we have
\begin{align}
     x_{t,i}   = \overline x_{\floor{t}_k + 1} - \alpha \sum_{l = \floor{t}_k +1 }^{t-1} \frac{m_{l,i}}{\sqrt{\hat v_l}}, \nonumber
\end{align}
since $x_{t,i}'s$ are averaged on steps $ck +1, c \in \mathbb N_0$.  Thus, we have
\begin{align}
     \left \| {\overline x_t - x_{t,i}}{} \right\|^2  = & \alpha ^2  \left \|   \sum_{l = \floor{t}_k +1 }^{t-1} \left( \frac{m_{l,i}}{\sqrt{\hat v_l}}  - \frac{1}{N}\sum_{o=1} ^ N \frac{m_{l,o}}{\sqrt{\hat v_o}}  \right) \right\|^2 \nonumber \\
    \leq & 4 N(k-1)^2 \alpha^2 d \frac{G^2}{\epsilon}, \nonumber
\end{align}
because $(t-1) -  (\floor{t}_k +1) +1\leq k-1$, and $(\hat v_o)_j \geq \epsilon,  \forall j \in [d], o \in [N]$, and $\|m_{t,o}\|_{\infty} \leq G, \forall t, o$. \hfill $\square$

\vspace{0.05in}

Combining~\eqref{eq: bias_bound},~\eqref{eq: z_x_lip},~\eqref{eq: x_x_avg_diff},~\eqref{eq: z_x_bound} and~\eqref{eq: consensus_bound}, we obtain
\begin{align}\notag
    \left \| \frac{\nabla f(\overline z_t) - \overline {\nabla f} (x_t)}{\hat v_t^{1/4}} \right\|^2 
    \leq & \frac{L}{\min_j (\hat v_t^{1/2})_j} \left(2 \frac{\beta_1^2}{(1-\beta_1)^2} +8(k-1)^2  \right) \alpha^2  d \frac{G^2}{\epsilon}.  \nonumber
\end{align}
which is the desired bound. This completes the proof. \hfill $\square$

\vspace{0.05in}

Therefore, by~\eqref{eq: bias_split} and~\eqref{eq: bias_overall_bound}, we have
\begin{align}\label{eq: bias_consensus_bounded}
    -\langle \nabla f(\overline z_t), \frac{ \overline {\nabla f} (x_t)}{\sqrt{\hat v_t}} \rangle
    \leq &  -  \frac{1}{2}\left \| \frac{\nabla f(\overline z_t)}{\hat v_t^{1/4}}\right\|^2 - \frac{1}{2}\left\| \frac{\overline {\nabla f} (x_t)}{\hat v_t^{1/4}} \right\|^2  + \alpha^2 \frac{L}{\min_j (\hat v_t^{1/2})_j} \left( \frac{\beta_1^2}{(1-\beta_1)^2} +4(k-1)^2  \right)   d \frac{G^2}{\epsilon}.
\end{align}
Substituting~\eqref{eq: bias_consensus_bounded} into~\eqref{eq: first_order} and~\eqref{eq: lipschitz_exp} yields
\begin{align}
    &\alpha \mathbb E \left[ \frac{1}{2}\left \| \frac{\nabla f(\overline z_t)}{\hat v_t^{1/4}}\right\|^2 + \frac{1}{2}\left\| \frac{\overline {\nabla f} (x_t)}{\hat v_t^{1/4}} \right\|^2\right]  \nonumber \\
    \leq  & \mathbb E[   f(\overline  z_{t})] -  \mathbb E [f(\overline  z_{t+1})] + \frac{L}{2 }\mathbb E[ \|\overline z_{t+1} -  \overline z_t \|^2]  \nonumber \\
    &\hspace{0.5in}+ \alpha \mathbb E \left [ \langle  \nabla f(\overline z_t), \frac{\beta_1}{1-\beta_1 } \left(\frac{1}{\sqrt{\hat v_{t-1}}} - \frac{1}{\sqrt{\hat v_{t}}}\right) \odot \overline m_{t-1}\right ]  \nonumber \\
    &\hspace{0.5in}+ \alpha^3 \mathbb E \left[ \frac{L}{\min_j (\hat v_t^{1/2})_j} \left( \frac{\beta_1^2}{(1-\beta_1)^2} +4(k-1)^2  \right)   d \frac{G^2}{\epsilon} \right]. \nonumber
\end{align}
Summing over $t$ from 1 to $T$ and divide both sides by $T\alpha$, we get
{
\begin{align}\label{eq: remain_t1_t2}
    &\frac{1}{T}  \sum_{t=1} ^T \mathbb E \left[ \frac{1}{2}\left \| \frac{\nabla f(\overline z_t)}{\hat v_t^{1/4}}\right\|^2 + \frac{1}{2}\left\| \frac{\overline {\nabla f} (x_t)}{\hat v_t^{1/4}} \right\|^2\right]  \nonumber \\
    \leq  & \frac{1}{T\alpha} (\mathbb E[   f(\overline  z_{1})] -  \mathbb E [f(\overline  z_{T+1})]) + \frac{L}{2 } \frac{1}{T\alpha} \underbrace{\sum_{t=1}^T \mathbb E[ \|\overline z_{t+1} - \overline z_t \|^2]}_{T_1}  \nonumber \\
    &+ \frac{1}{T}\underbrace{\sum_{t=1} ^T \mathbb E \left [ \left \langle  \nabla f(\overline z_t), \frac{\beta_1}{1-\beta_1 } \left(\frac{1}{\sqrt{\hat v_{t-1}}} - \frac{1}{\sqrt{\hat v_{t}}}\right) \odot \overline m_{t-1}\right\rangle \right ]}_{T_2}  \nonumber \\
    &+  \alpha^2 \frac{1}{T}\sum_{t=1}^T \mathbb E \left[ \frac{L}{\min_j (\hat v_t^{1/2})_j} \left( \frac{\beta_1^2}{(1-\beta_1)^2} +4(k-1)^2  \right)   d \frac{G^2}{\epsilon} \right].
\end{align}
}

What remains is to bounded $T_1$ and $T_2$.

\newpage

\begin{lemma} We have
\begin{align}
    T_1 \leq & 2\alpha^2 \sum_{t=1}^T \mathbb E \left [\left\| \frac{\overline {\nabla f}  (x_{t})}{\sqrt{\hat v_t}}\right\|^2 +  \frac{1}{N}   \sigma^2  \left\| \frac{ 1}{\sqrt{\hat v_t}} \right\|^2   \right]+  2\alpha^2  \frac{\beta_1^2}{(1-\beta_1)^2 } G^2 \frac{d}{\epsilon}. \nonumber
\end{align}
\end{lemma}
\noindent\textbf{Proof:}
 By~\eqref{eq: z_update}, we know that
\begin{align}
    & \|\overline z_{t+1} - \overline z_t \|^2  \nonumber \\
    =& \alpha^2 \left \| \frac{\beta_1}{1-\beta_1 } \left(\frac{1}{\sqrt{\hat v_{t-1}}} - \frac{1}{\sqrt{\hat v_{t}}}\right) \odot \overline m_{t-1} - \alpha \frac{\overline g_t}{\sqrt{\hat v_t}}\right\|^2 \nonumber \\
    \leq & 2\alpha^2 \left(\left \| \frac{\beta_1}{1-\beta_1 } \left(\frac{1}{\sqrt{\hat v_{t-1}}} - \frac{1}{\sqrt{\hat v_{t}}}\right) \odot \overline m_{t-1} \right\|^2 +  \left\|  \frac{\overline g_t}{\sqrt{\hat v_t}}\right\|^2\right). \nonumber
\end{align}

In addition, conditioned on all randomness in $\hat v_t$ (gradients from iteration $1$ until $t-1$), we have
\begin{align}
    \mathbb E \left[  \left\|  \frac{\overline g_t}{\sqrt{\hat v_t}}\right\|^2\right] =& \mathbb E \left[  \left\| \frac{1}{N}\sum_{i=1}^N \frac{ g_{t,i}}{\sqrt{\hat v_t}}\right\|^2\right] = \frac{1}{N^2}  \mathbb E \left[ \sum_{j=1}^N \sum_{i=1}^N  \left \langle \frac{ g_{t,i}}{\sqrt{\hat v_t}},\frac{ g_{t,j}}{\sqrt{\hat v_t}} \right\rangle \right]  \nonumber \\
    \stackrel{(a)}{=} &  \frac{1}{N^2}  \mathbb E \left[ \sum_{j=1}^N \sum_{i=1}^N  \left \langle \frac{ \nabla f_i(x_{t,i}) + \xi_{t,i}}{\sqrt{\hat v_t}},\frac{\nabla f_j(x_{t,j}) + \xi_{t,j}}{\sqrt{\hat v_t}} \right\rangle \right] \nonumber \\
    \stackrel{(b)}{=} &  \frac{1}{N^2} \left\| \frac{ \sum_{i=1}^N \nabla f_i(x_{t,i})}{\sqrt{\hat v_t}}\right\|^2 +  \frac{1}{N^2} \sum_{i=1}^N   \mathbb E \left[ \left\| \frac{ \xi_{t,i}}{\sqrt{\hat v_t}} \right\|^2  \right] \nonumber \\
    \stackrel{(c)}{\leq} & \left\| \frac{\overline {\nabla f}  (x_{t})}{\sqrt{\hat v_t}}\right\|^2 +  \frac{1}{N}   \sigma^2  \left\| \frac{ 1}{\sqrt{\hat v_t}} \right\|^2, \nonumber
\end{align}
where $(a)$ is a reparameterization of the noises on gradients, $(b)$ is due to the assumption that all elements of $\xi_{t,i}$'s are independent of each other and $\mathbb E[(\xi_{t,i})_j] = 0, \forall j \in [d]$, and (c) is by the assumption that $ E[(\xi_{t,i})_j^2] \leq \sigma$ and the fact that  $\hat v_t$ and $x_{t,i}$ are both fixed given all gradients before iteration $t$. Furthermore, we have
\begin{align}
    & \left \| \frac{\beta_1}{1-\beta_1 } \left(\frac{1}{\sqrt{\hat v_{t-1}}} - \frac{1}{\sqrt{\hat v_{t}}}\right) \odot \overline m_{t-1} \right\|^2 \nonumber \\
    \leq & \frac{\beta_1^2}{(1-\beta_1)^2 } G^2 \sum_{j=1}^d  \left(\frac{1}{\sqrt{(\hat v_{t-1})_j}} - \frac{1}{\sqrt{(\hat v_{t})_j}} \right)^2 \nonumber \\
    \leq &\frac{\beta_1^2}{(1-\beta_1)^2 } G^2 \frac{1}{\epsilon^{1/2}}  \sum_{j=1}^d
\left|\frac{1}{\sqrt{(\hat v_{t-1})_j}} - \frac{1}{\sqrt{(\hat v_{t})_j}}, \right| \nonumber
\end{align}
and thus
\begin{align}
    & \sum_{t=1}^T \left \| \frac{\beta_1}{1-\beta_1 } \left(\frac{1}{\sqrt{\hat v_{t-1}}} - \frac{1}{\sqrt{\hat v_{t}}}\right) \odot \overline m_{t-1} \right\|^2 \nonumber \\
    \leq &\sum_{t=1}^T \frac{\beta_1^2}{(1-\beta_1)^2 } G^2 \frac{1}{\epsilon^{1/2}}  \sum_{j=1}^d
\left|\frac{1}{\sqrt{(\hat v_{t-1})_j}} - \frac{1}{\sqrt{(\hat v_{t})_j}} \right| \nonumber \\
\leq & \frac{\beta_1^2}{(1-\beta_1)^2 } G^2 \frac{1}{\epsilon^{1/2}}  \sum_{j=1}^d
\left|\frac{1}{\sqrt{(\hat v_{0})_j}}\right|  =  \frac{\beta_1^2}{(1-\beta_1)^2 } G^2 \frac{1}{\epsilon} d,  \nonumber
\end{align}
where the last inequality is due to non-decreasing property of $\hat v_t$.

Combining all above, we obtain
\begin{align}\notag
    T_1 \leq & 2\alpha^2 \sum_{t=1}^T \mathbb E \left [\left\| \frac{\overline {\nabla f}  (x_{t})}{\sqrt{\hat v_t}}\right\|^2 +  \frac{1}{N}   \sigma^2  \left\| \frac{ 1}{\sqrt{\hat v_t}} \right\|^2   \right]
    +  2\alpha^2  \frac{\beta_1^2}{(1-\beta_1)^2 } G^2 \frac{d}{\epsilon}.
\end{align}
This completes the proof.\hfill $\square$

\vspace{0.1in}

Now we can examine $T_2$. We have
\begin{align}
    T_2 = & \sum_{t=1} ^T \mathbb E \left [ \left \langle  \nabla f(\overline z_t), \frac{\beta_1}{1-\beta_1 } \left(\frac{1}{\sqrt{\hat v_{t-1}}} - \frac{1}{\sqrt{\hat v_{t}}}\right) \odot \overline m_{t-1}\right \rangle \right ] \nonumber \\
    \leq &  \mathbb E \left [\sum_{t=1} ^T \frac{\beta_1}{1-\beta_1 } G^2   \sum_{j=1}^d \left|\frac{1}{\sqrt{(\hat v_{t-1})_j}} - \frac{1}{\sqrt{(\hat v_{t})_j}}\right| \right ] \nonumber \\
    \leq &  \mathbb E \left [ \frac{\beta_1}{1-\beta_1 } G^2   \sum_{j=1}^d \left|\frac{1}{\sqrt{(\hat v_{0})_j}}\right| \right ] = \frac{\beta_1}{1-\beta_1 } G^2   d\frac{1}{\epsilon^{1/2}},   \nonumber
\end{align}
where the second inequality is because $\hat v_t$ is non-decreasing.

Substituting the bounds on $T_1$ and $T_2$ into~\eqref{eq: remain_t1_t2}, we obtain
\begin{align}\notag
   &\frac{1}{T}  \sum_{t=1} ^T \mathbb E \left[ \frac{1}{2}\left \| \frac{\nabla f(\overline z_t)}{\hat v_t^{1/4}}\right\|^2 + \frac{1}{2}\left\| \frac{\overline {\nabla f} (x_t)}{\hat v_t^{1/4}} \right\|^2\right] \\\notag
    \leq  & \frac{1}{T\alpha} (\mathbb E[   f(\overline  z_{1})] -  \mathbb E [f(\overline  z_{T+1})]) \nonumber \\
    & +  \frac{L}{T} \alpha \sum_{t=1}^T \mathbb E \left [\left\| \frac{\overline {\nabla f}  (x_{t})}{\sqrt{\hat v_t}}\right\|^2 +  \frac{1}{N}   \sigma^2  \left\| \frac{ 1}{\sqrt{\hat v_t}} \right\|^2 \right]  \nonumber \\
    &+ \frac{1}{T}\frac{\beta_1}{1-\beta_1 } G^2   d\frac{1}{\epsilon^{1/2}} + \frac{L}{T}\alpha^2  \frac{\beta_1^2}{(1-\beta_1)^2 } G^2 \frac{d}{\epsilon}  \nonumber \\\label{eq: semi_final}
    &+  \alpha^2 \frac{1}{T}\sum_{t=1}^T \mathbb E \left[ \frac{L}{\min_j (\hat v_t^{1/2})_j} \left( \frac{\beta_1^2}{(1-\beta_1)^2} +4(k-1)^2  \right)   d \frac{G^2}{\epsilon} \right].
\end{align}

\newpage

Further, by choosing $\alpha = \min(\frac{\sqrt{N}}{\sqrt{Td}},\frac{\sqrt{\epsilon}}{4L})$, we have
\begin{align}\label{eq: cancel_descent}
    &{L} \frac{1}{T} \alpha \sum_{t=1}^T \mathbb E \left [\left\| \frac{\overline {\nabla f}  (x_{t})}{\sqrt{\hat v_t}}\right\|^2  \right]
    \leq  {L} \frac{1}{T} \alpha \sum_{t=1}^T \mathbb E \left [\frac{1}{\sqrt{\epsilon}}\left\| \frac{\overline {\nabla f}  (x_{t})}{{\hat v_t^{1/4}}}\right\|^2  \right]
     \frac{1}{4} \frac{1}{T}  \sum_{t=1}^T \mathbb E \left [\left\| \frac{\overline {\nabla f}  (x_{t})}{{\hat v_t^{1/4}}}\right\|^2  \right].
\end{align}

Thus, from~\eqref{eq: semi_final} and~\eqref{eq: cancel_descent}, we obtain
\begin{align}\notag
   &\frac{1}{T}  \sum_{t=1} ^T \mathbb E \left[ \frac{1}{2}\left \| \frac{\nabla f(\overline z_t)}{\hat v_t^{1/4}}\right\|^2 + \frac{1}{4}\left\| \frac{\overline {\nabla f} (x_t)}{\hat v_t^{1/4}} \right\|^2\right] \\\notag
    \leq  & \frac{1}{T\alpha} (\mathbb E[   f(\overline  z_{1})] -  \mathbb E [f(\overline  z_{T+1})]) + {L}  \frac{\sqrt{d}}{\sqrt{TN}}  \sigma^2  \frac{1}{\epsilon} \nonumber \\
    &   + \frac{1}{T}\frac{\beta_1}{1-\beta_1 } G^2   d\frac{1}{\epsilon^{1/2}}   + \frac{LN}{T^2} \frac{\beta_1^2}{(1-\beta_1)^2 } \frac{G^2 }{\epsilon}   \nonumber \\
    &+  \frac{N}{T}   {L} \left( \frac{\beta_1^2}{(1-\beta_1)^2} +4(k-1)^2  \right)    \frac{G^2}{\epsilon^{1.5}}, \nonumber
\end{align}
and when $T\geq \frac{16NL^2}{\epsilon d}$, we have
\begin{align}\notag
   &\frac{1}{T}  \sum_{t=1} ^T \mathbb E \left[ \frac{1}{2}\left \| \frac{\nabla f(\overline z_t)}{\hat v_t^{1/4}}\right\|^2 + \frac{1}{4}\left\| \frac{\overline {\nabla f} (x_t)}{\hat v_t^{1/4}} \right\|^2\right]\\\notag
    \leq  & \frac{\sqrt{d}}{\sqrt{TN}} (\mathbb E[   f(\overline  z_{1})] -  \mathbb E [f(\overline  z_{T+1})]) + {L}  \frac{\sqrt{d}}{\sqrt{TN}}  \sigma^2  \frac{1}{\epsilon}  \nonumber \\
    &  + \frac{d}{T}\frac{\beta_1}{1-\beta_1 } G^2   \frac{1}{\epsilon^{1/2}}  + \frac{LN}{T^2} \frac{\beta_1^2}{(1-\beta_1)^2 } \frac{G^2 }{\epsilon}  \nonumber \\\label{eq: final3}
    &+  \frac{N}{T}   {L} \left( \frac{\beta_1^2}{(1-\beta_1)^2} +4(k-1)^2  \right)    \frac{G^2}{\epsilon^{1.5}}.
\end{align}

At this point, we have obtained the $O(\frac{\sqrt{d}}{\sqrt{TN}})$ convergence rate (when $T$ is sufficiently large), which matches the convergence rate of SGD.  One remaining item is to convert the convergence measure to the norm of gradients of $f$. We do this by the following Lemma.

\vspace{0.1in}

\begin{lemma} \label{lem: convert_measure}
For Algorithm~\ref{alg: local_ams}, we have
{
\begin{align}
     &  \left\| \frac{\overline {\nabla f} (x_t)}{\hat v_t^{1/4}} \right\|^2
     \geq  \frac{1}{2}    \left\| \frac{ {\nabla f} (\overline x_t)}{\hat v_t^{1/4}} \right\|^2 -4L(k-1)^2 \alpha^2 d \frac{G^2}{\epsilon^{1.5}}.  \nonumber
\end{align}
}%
\end{lemma}
\noindent\textbf{Proof}: We have
\begin{align} \label{eq: descent_split}
     \left\| \frac{\overline {\nabla f} (x_t)}{\hat v_t^{1/4}} \right\|^2
     \geq &  \frac{1}{2}\left\| \frac{ {\nabla f} (\overline x_t)}{\hat v_t^{1/4}} \right\|^2- \left\| \frac{\overline {\nabla f} (x_t) - {\nabla f} (\overline x_t)}{\hat v_t^{1/4}} \right\|^2 \nonumber \\
     \geq &  \frac{1}{2}\left\| \frac{ {\nabla f} (\overline x_t)}{\hat v_t^{1/4}} \right\|^2 -  \frac{1}{N} \sum_{i=1}^N \left\|  \frac{ {\nabla f_i} (x_{t,i}) - {\nabla f_i} (\overline x_t)}{\hat v_t^{1/4}} \right\|^2 \nonumber \\
     \geq & \frac{1}{2}\left\| \frac{ {\nabla f} (\overline x_t)}{\hat v_t^{1/4}} \right\|^2 - 4L(k-1)^2 \alpha^2 d \frac{G^2}{\epsilon^{1.5}},  \nonumber
\end{align}
where the first inequality is due to Cauchy-Schwartz, the second inequality is due to Jensen's inequality, and the last inequality is due to Lemma~\ref{lem: z_x_diff} and L-smoothness of $f_i$ (A1). \hfill $\square$

\vspace{0.1in}

We can then transform~\eqref{eq: final3} into
\begin{align}
    \frac{1}{T}  \sum_{t=1} ^T \mathbb E \left[ \frac{1}{8}\left\| \frac{ {\nabla f} (\overline x_t)}{\hat v_t^{1/4}} \right\|^2\right]  
    \leq  & \frac{\sqrt{d}}{\sqrt{TN}} (\mathbb E[   f(\overline  z_{1})] -  \mathbb E [f(\overline  z_{T+1})]) + {L}  \frac{\sqrt{d}}{\sqrt{TN}}  \sigma^2  \frac{1}{\epsilon}\nonumber \\
    &   + \frac{d}{T}\frac{\beta_1}{1-\beta_1 } G^2   \frac{1}{\epsilon^{1/2}} +   \frac{LN}{T^2} \frac{\beta_1^2}{(1-\beta_1)^2 } \frac{G^2 }{\epsilon} \nonumber  \\
    &+  \frac{N}{T}   {L} \left( \frac{\beta_1^2}{(1-\beta_1)^2} +5(k-1)^2  \right)    \frac{G^2}{\epsilon^{1.5}}, \nonumber
\end{align}
by Lemma~\ref{lem: convert_measure}. Multiplying both sides of the above  inequality by 8 and using the fact that $z_1 = x_1$ complete the overall proof.
\hfill $\square$

\vspace{0.5in}

\bibliography{ref}
\bibliographystyle{plainnat}

\end{document}